\documentclass{article}

\usepackage{microtype}
\usepackage{graphicx}
\usepackage{subfigure}
\usepackage{booktabs} 

\usepackage{hyperref}

\usepackage[accepted]{icml2019}

\icmltitlerunning{Complementary-Label Learning for Arbitrary Losses and Models}

\usepackage{amsmath,amsthm,amsfonts,amssymb}
\usepackage{bm}
\usepackage{comment}
\usepackage{tablefootnote}
\usepackage{multirow}
\usepackage{hhline}
\usepackage{graphicx}

\usepackage{algorithm,algorithmic}

\usepackage[font=small,labelfont=bf]{caption}
\usepackage{url}

\newcommand{\bmI}{\bm{I}}\newcommand{\bmT}{\bm{T}}\newcommand{\bmg}{\bm{g}}

\newtheorem{theorem}{Theorem}

\newtheorem{corollary}[theorem]{Corollary}

\begin{document}

\twocolumn[
\icmltitle{Complementary-Label Learning for Arbitrary Losses and Models}




\begin{icmlauthorlist}
\icmlauthor{Takashi Ishida}{utokyo,riken}
\icmlauthor{Gang Niu}{riken}
\icmlauthor{Aditya Krishna Menon}{google}
\icmlauthor{Masashi Sugiyama}{riken,utokyo}
\end{icmlauthorlist}

\icmlaffiliation{utokyo}{The University of Tokyo}
\icmlaffiliation{riken}{RIKEN}
\icmlaffiliation{google}{Google Research}

\icmlcorrespondingauthor{Takashi Ishida}{ishida@ms.k.u-tokyo.ac.jp}

\icmlkeywords{complementary labels, weakly-supervised learning}

\vskip 0.3in
]



\printAffiliationsAndNotice{}  

\begin{abstract}
In contrast to the standard classification paradigm where
the true class is given to each training pattern,
\emph{complementary-label learning} only uses training patterns each equipped
with a complementary label,
which only specifies one of the classes that the pattern does \emph{not} belong to.
The goal of this paper is to derive a novel framework of complementary-label learning with an unbiased estimator of the classification risk, for arbitrary losses and models---all existing methods have failed to achieve this goal.
Not only is this beneficial for the learning stage, it also makes model/hyper-parameter selection (through cross-validation) possible without the need of any ordinarily labeled validation data, while using any linear/non-linear models or convex/non-convex loss functions.
We further improve the risk estimator by a non-negative correction and gradient ascent trick, and demonstrate its superiority through experiments.
\end{abstract}

\section{Introduction}
Modern classification methods usually require massive data with high-quality labels, but preparing such datasets is unrealistic in many domains.
To mitigate the problem, previous works have investigated ways to learn from weak supervision: \emph{semi-supervised learning} \citep{chapelle06SSL,miyato16iclr,kipf17iclr,sakai17icml,tarvainen17nips,oliver18neurips}, \emph{noisy-label learning} \citep{natarajan13nips,menon15icml,patrini17cvpr,ma18icml,han18neurips,charoenphakdee19icml}, \emph{positive-unlabeled learning} \citep{elkan08kdd,christo14nips,kiryo17nips}, \emph{positive-confidence learning} \citep{ishida18nips}, \emph{similar-unlabeled learning} \citep{bao18icml}, \emph{unlabeled-unlabeled learning} \citep{duPlessis13taai,nan19iclr}, and others.

In this paper, we consider learning from another natural type of weak supervision called \emph{complementary-label learning} \citep{ishida17nips,yu17eccv}, where the label only specifies one of the classes that the pattern does \emph{not} belong to.  For example, a crowdsourced worker can tell us a pattern does not belong to a certain class, instead of identifying the correct class.
In contrast to the ordinary case where the true class is given to each pattern (which often needs to be chosen out of many candidate classes precisely), collecting these complementary labels is obviously much easier and less costly.

Another potential application is collecting survey data that requires extremely private questions \citep{ishida17nips}.  It would be less mentally demanding, if we explain to the respondent that we will transform their provided true label to a complementary label, before the data is saved into the database. This might become common in the future where privacy concerns are increasing.

A natural question is, however, is it possible to learn from such complementary labels (without \emph{any} true labels)?

The problem has previously been tackled by \citet{ishida17nips},
showing that the classification risk can be recovered only from complementarily labeled data.
They also gave theoretical analysis with a statistical consistency guarantee.
However, they required strong restrictions on the loss functions, allowing only one-versus-all and pairwise comparison multi-class loss functions \citep{ova}, with certain non-convex binary losses.
This is a severe limitation since the softmax cross-entropy loss, which cannot be expressed by the two losses above, is the most popular loss in deep learning nowadays.

Later, \citet{yu17eccv} proposed a different formulation for complementary labels by employing the forward loss correction technique \citep{patrini17cvpr} to adjust the learning objective, but limiting the loss function to softmax cross-entropy loss.
Their proposed risk estimator is not necessarily \emph{unbiased} but the minimizer is theoretically guaranteed to be \emph{consistent} with the minimizer of the risk for ordinary labels (under an implicit assumption on the model for convergence analysis).
They also extended the problem setting to where complementary labels are chosen in an uneven (biased) way.

In this paper, we first derive an unbiased risk estimator with a general loss function, making \emph{any} loss functions available for use: not only the softmax cross-entropy loss function but other convex/non-convex loss functions can also be applied. We also do not have implicit assumptions on the classifier, allowing both linear and non-linear models.
We also prove that our new framework is a generalization of previous complementary-label learning \citep{ishida17nips}.

\citet{yu17eccv} does not have an unbiased risk estimator, which means users will need clean data with true labels to calculate the error rate during the validation process.
On the other hand, our proposed unbiased risk estimator can handle \emph{complementarily} labeled validation data not only for our learning objective, but also for that of \citet{yu17eccv}.
This is helpful since collecting clean data is usually much more expensive.  Note that in the example of survey with extremely private questions explained earlier, it may be impossible to even collect a small number of validation data with true labels.

Finally, our proposed unbiased risk estimator has an issue that the classification risk can attain negative values after learning, leading to overfitting.
We further propose a non-negative correction to the original unbiased risk estimator to improve our estimator.
The modified objective is no longer guaranteed to be an unbiased risk estimator, but the unbiased risk estimator can still be used for validation procedures for this modified learning objective.
We experimentally show that our proposed method is comparable to or better than previous methods \citep{ishida17nips,yu17eccv} in terms of classification accuracy.

A summary of our contributions is as follows:
\begin{itemize}
  \item We propose a new unbiased risk estimator, allowing usage of any loss (convex, non-convex) and any model (parametric, non-parametric) for complementary-label learning.
  \item This risk can be used not only as a learning objective, but as a validation criterion even for other methods, such as \citet{ishida17nips} and \citet{yu17eccv}.
  \item We further investigate correction schemes to make complementary-label learning practical and demonstrated the performance in experiments.
\end{itemize}

\begin{table*}[t]
\center
  \caption{Comparison of two proposed complementary-label methods with previous works.  We first propose a general unbiased risk estimator for complementary labels that has no restrictions on loss functions and models.  We next propose a modified non-negative formulation which solves overfitting issues and leads to better experimental results. Even though the non-negative formulation is no longer an unbiased estimator as a learning objective, the unbiased estimator can be used in the validation procedure.}
  \label{tb:compareproposed}
  \begin{tabular}{@{\ }l|ccccc@{\ }}
\toprule
\multirow{2}{*}{Methods} & \multirow{2}{*}{\shortstack{loss assump.\\free}} & \multirow{2}{*}{\shortstack{model assump.\\free}} & \multirow{2}{*}{\shortstack{unbiased\\estimator}} &  \multirow{2}{*}{\shortstack{explicit risk\\correction}} \\
&&&\\\midrule
\citet{ishida17nips} & $\times$ & $\checkmark$ &$\checkmark$& $\times$\\
\citet{yu17eccv} & $\times$ & $\times$ &$\times$&$\times$\\\midrule
Proposed (General formulation) & $\checkmark$ & $\checkmark$ &$\checkmark$&$\times$\\
Proposed (Non-negative formulation) & $\checkmark$ & $\checkmark$ &$\times$ &$\checkmark$\\
\bottomrule
  \end{tabular}
\end{table*}

\section{Review of previous works}
\label{sc:review}
In this section, we introduce some notations and review the formulations of learning from ordinary labels, learning from complementary labels, learning from ordinary \& complementary labels, and learning from partial labels.
\subsection{Learning from ordinary labels}
Let $\mathcal{X}$ be an instance space and $\mathcal{D}$ be the joint distribution over $\mathcal{X}\times [K]$ for class label set $[K]:=\{1,2,\ldots, K\}$, with random variables $(X,Y) \sim \mathcal{D}$.
The data at hand is sampled independently and identically from the joint distribution: $\{(x_i, y_i)\}^n_{i=1}\overset{\text{i.i.d.}}{\sim} \mathcal{D}$.
The joint distribution $\mathcal{D}$ can be either decomposed into class-conditionals $\{P_k\}^K_{k=1}$ and base rate $\{\pi_k\}^K_{k=1}$, where $P_k := \mathbb{P}(X|Y=k)$ and $\pi_k:=\mathbb{P}(Y=k)$, or the marginal $M$ and class-probability function $\bm{\eta}:\mathcal{X}\rightarrow \Delta_k$, where $M:=\mathbb{P}(X)$, $\bm{\eta}_k(x):=\mathbb{P}(Y=k|X=x)$ and $\Delta_K$ is the conditional probability simplex for $K$ classes.
A loss is any $\ell:[K]\times \mathbb{R}^K\rightarrow \mathbb{R}_+$. The decision function is any $\bmg:\mathcal{X}\rightarrow\mathbb{R}^K$ and $\bmg_k(X)$ is the $k$-th element of $\bmg(X)$.
The risk for the decision function $\bmg$ with respect to loss $\ell$ and implicit distribution $\mathcal{D}$ is:
\begin{align}
\label{risk:ordinary}
  R(\bmg;\ell):&= \mathbb{E}_{(X,Y)\sim \mathcal{D}}[\ell(Y,\bmg(X))],
\end{align}
where $\mathbb{E}$ denotes the expectation.  Two useful equivalent expressions of classification risk \eqref{risk:ordinary} used in later sections are
\begin{align}
\label{risk:class_prior_decomposition}
  R(\bmg;\ell)=& \mathbb{E}_X[\bm{\eta}(x)^\top\bm{\ell}(\bmg(X))]
                     = \sum_{k=1}^K \pi_k \mathbb{E}_{\mathbb{P}_k}\Big[\ell(k,\bmg(X))\Big],
\end{align}
where,
\begin{align*}
\bm{\ell}(\bmg(X)) := [\ell(1,\bmg(X)), \ell(2,\bmg(X)), \ldots, \ell(K, \bmg(X))]^\top.
\end{align*}
The goal of classification is to learn the decision function $\bmg$ that minimizes the risk.  In the usual classification case with ordinarily labeled data at hand, approximating the risk empirically is straightforward:
\begin{align*}
\widehat{R}(\bmg;\ell) := \frac{1}{n}\sum^n_{i=1}\ell(y_i, \bmg(x_i)).
\end{align*}
Some well known multi-class loss functions are one-versus-all and pairwise comparison losses:
\begin{align}
\ell_{\text{OVA}}\big(k,\bmg(x)\big) &= s\big(\bmg_k(x)\big)+\frac{1}{K-1}\sum_{k'\neq k}s\big(-\bmg_{k'}(x)\big),\\
\ell_{\text{PC}}\big(k,\bmg(x)\big) &= \sum_{k'\neq k}s\big(\bmg_k(x)-\bmg_{k'}(x)\big),
\end{align}
where $s(z):\mathbb{R}\rightarrow\mathbb{R}_+$ is a binary loss function.
\subsection{Learning from complementary labels}
Next we consider the problem of learning from complementary labels \citep{ishida17nips}.  We observe patterns each equipped with a complementary label $\{(x_{i'},\overline{y}_{i'})\}^{n'}_{i'=1}$ sampled independently and identically from a different joint distribution $\overline{\mathcal{D}}\neq\mathcal{D}$.  We denote random variables as $(X, \overline{Y}) \sim \overline{\mathcal{D}}$.
As before, we assume this distribution can be decomposed into either class-conditionals $\{\overline{P}_k\}^K_{k=1}$ and base rate $\{\overline{\pi}\}^K_{k=1}$, or marginal $M$ and class-probability function $\bm{\overline{\eta}}:\mathcal{X}\rightarrow\Delta_K$, where $\overline{P}_k:=\mathbb{P}(X|\overline{Y}=k)$, $\overline{\pi}_k := \mathbb{P}(\overline{Y}=k)$, $M:=\mathbb{P}(X)$, $\bm{\overline{\eta}}_k(x):=\mathbb{P}(\overline{Y}=k|X=x)$, and $\overline{Y}$ is the complementary label.

Without any assumptions on $\overline{\mathcal{D}}$, it is impossible to design a suitable learning procedure.
The assumption for unbiased complementary learning used in \citet{ishida17nips} was
\begin{align}
\label{unbiased-complementary-label-assumption}
  \bm{\overline{\eta}}(x) = \bmT \bm{\eta}(x),
\end{align}

where $\bmT \in \mathbb{R}^{K\times K}$ is a matrix that takes $0$ on diagonals and $\frac{1}{K-1}$ on non-diagonals.

This assumption implies all other labels are chosen with uniform probability.  This can be forced by designing the data collecting system to first pick up a label randomly and then ask the worker if the data belong to the label with a yes or no.  When the answer is no, we will attach that label as the complementary label, and the data will follow the uniform assumption.
Under this assumption, \citet{ishida17nips} proved that they can recover the classification risk \eqref{risk:ordinary} from an alternative formulation using only complementarily labeled data when the loss function satisfies certain conditions.
More specifically, usable loss functions are one-versus-all or pairwise comparison multi-class loss functions \citep{ova}:
\begin{align}
\overline{\ell}_{\text{OVA}}\big(\overline{k},\bmg(x)\big) &= \frac{1}{K-1}\sum_{k\neq \overline{k}}s\big(\bmg_k(x)\big)+s\big(-\bmg_{\overline{k}}(x)\big) \label{comp-ova}\\
\overline{\ell}_{\text{PC}}\big(\overline{k},\bmg(x)\big) &= \sum_{k'\neq \overline{k}}s\big(\bmg_k(x)-\bmg_{\overline{k}}(x)\big)\label{comp-pc}
\end{align}
each with binary loss function $s(z)$ that satisfies $s(z)+s(-z) = 1$, such as ramp loss $s_\text{R}(z) = \frac{1}{2}\max\big(0, \min(2,1-z)\big)$ or sigmoid loss $s_\text{S}(z) = \frac{1}{1+e^z}$.

Having an unbiased risk estimator is also helpful for the validation process.  Since we do not have ordinary labels in our validation set in the complementary-label learning setting, we cannot follow the usual validation procedure that uses zero-one error or accuracy.  If we have an unbiased estimator of the original classification risk (which can be interpreted as zero-one error), we can use the empirical risk for (cross)-validated complementary data to select the best hyper-parameter or deploy early stopping.

An extension of the above method was considered in \citet{yu17eccv} by using a different assumption than \eqref{unbiased-complementary-label-assumption}:
there is some bias amongst the possible complementary labels that can be chosen, thus the non-diagonals of $\bmT$ is not restricted to $\frac{1}{K-1}$.
However, one will need to estimate $\bmT$ beforehand, which is fairly difficult without strong assumptions.
Furthermore, in this setup, it is necessary to encourage the worker to provide more difficult complementary labels, for example, by giving higher rewards to certain classes.  Otherwise, the complementary label given by the worker may be too obvious and uninformative.  Even though the two assumptions are mathematically similar, the data generation process may be different.  In this paper we focus on the former assumption.

Unlike \citet{ishida17nips}, \citet{yu17eccv} did not directly provide a risk estimator, but they showed that the \emph{minimizer} of their learning objective agrees with the minimizer of the original classification risk \eqref{risk:ordinary}.
Note that, in their formulation, the loss function is restricted to the softmax cross-entropy loss. Furthermore, the use of a highly non-linear model is supposed for consistency guarantee in their theoretical analysis.
Since the learning objective of \citet{yu17eccv} does not correspond to the classification risk, one will need clean data with true labels to calculate the error rate during the validation process.
On the other hand, our proposed risk estimator in this paper can cope with \emph{complementarily} labeled validation data not only for our own learning objective, but can be used to select hyper-parameters for others such as \citet{yu17eccv}.

\subsection{Learning from ordinary \& complementary labels}
In many practical situations, we may also have ordinarily labeled data in addition to complementarily labeled data.  \citet{ishida17nips} touched on the idea of crowdsourcing for an application with both types of data.
For example, we may choose one of the classes randomly by following the uniform distribution, with probability $\frac{1}{K-1}$ for each class, and ask crowdworkers whether a pattern belongs to the chosen class or not.
Then the pattern is treated as ordinarily labeled if the answer is yes; otherwise, the pattern is regarded as complementarily labeled.
If the true label was $y$ for a pattern, we can naturally assume that the crowdworker will answer yes by 
$\mathbb{P}(Y=y|X=x)$ and no by $1-\mathbb{P}(Y=y|X=x)$.
This way, ordinarily labeled data can be regarded as patterns from $\mathcal{D}$, and complementarily labeled data from $\overline{\mathcal{D}}$, justifying the assumption of unbiased complementary learning \eqref{unbiased-complementary-label-assumption}.

In \citet{ishida17nips}, they considered a convex combination of the classification risks derived from ordinarily labeled data and complementarily labeled data:
\begin{align*}
\alpha \overline{R}(\bmg;\overline{\ell}) + (1-\alpha) R(\bmg;\ell),
\end{align*}
where $\overline{R}(\bmg; \overline{\ell})= \mathbb{E}_{(X,\overline{Y})\sim\overline{\mathcal{D}}}[\overline{\ell}(\overline{Y},\bmg(X))]$ and $\alpha \in [0,1]$ is a hyper-parameter that interpolates between the two risks.
The combined (also unbiased) risk estimator can utilize both kinds of data in order to obtain better classifiers, which was demonstrated to perform well in experiments.

\subsection{Learning from partial labels}
In \emph{learning from partial labels} \citep{partial}, a candidate set of labels (which includes the correct class) is given to each pattern.  A different way to view complementary label is a candidate set that includes every class except the complementary label.  Even though the proposed method of \citet{partial} shows statistical consistency, it does not give an unbiased estimator of the classification risk.  Further, it has different assumptions, e.g., dominance relation, while \citet{ishida17nips} and this paper focus on  assumption \eqref{unbiased-complementary-label-assumption} with different data generation process and applications.

\section{Proposed method}\label{sc:proposed_method}
As discussed in the previous section, the method by \citet{ishida17nips} works well in practice, but it has restriction on the loss functions---the popular softmax cross-entropy loss is not allowed.
On the other hand, the method by \citet{yu17eccv} allows us to use the softmax cross-entropy loss, but it does not directly provide an estimator of the classification risk and thus model selection is problematic in practice.

We first describe our general unbiased risk formulation in Section \ref{sc:general_risk_formulation}.
Then we discuss how the estimator can be further improved in Section \ref{sec:necessity}.
Thirdly, we propose a way for our risk estimator to avoid overfitting by a \emph{non-negative risk estimator} in Section \ref{sc:nn_risk_estimator}.
Finally, we show practical implementation of our risk estimator with stochastic optimization methods in Section \ref{sc:implementation}.

\subsection{General risk formulation}
\label{sc:general_risk_formulation}
First, we describe our general unbiased risk formulation.
We give the following theorem, which allows unbiased estimation of the classification risk from complementarily labeled patterns:
\begin{theorem}\label{theorem:R(f)}
For any ordinary distribution $\mathcal{D}$ and complementary distribution $\overline{\mathcal{D}}$ related by \eqref{unbiased-complementary-label-assumption} with decision function $\bmg$, and loss $\ell$, we have
\begin{align}
R(\bmg; \ell) = \overline{R}(\bmg; \overline{\ell}) = \mathbb{E}_{(X,\overline{Y})\sim\overline{\mathcal{D}}}[\overline{\ell}(\overline{Y},\bmg(X))],
\end{align}
for the complementary loss
\begin{align}
\label{complementary_loss}
\overline{\bm{\ell}}\big(\bmg(x)\big):= \Big(-(K-1)\bmI_K+\bm{1}\bm{1}^\top\Big)\cdot \bm{\ell}\big(\bmg(x)\big),
\end{align}
or equivalently,
\begin{align}\label{scalar_complementary_loss}
  \overline{\ell}\big(k,\bmg(x)\big) = -(K-1)\cdot \ell\big(k,\bmg(x)\big) + \sum^K_{j=1}\ell\big(j,\bmg(x)\big),
\end{align}
where
$\bmI_K$ is a $K\times K$ identity matrix and $\bm{1}$ is a $K$-dimensional column vector with $1$ in each element.
\end{theorem}
Proof can be found in Appendix \ref{sec:theorem1proof}.
It is worth noting that, in the above derivation,
there are no constraints on the loss
function and classifier.
Thus, we can use any loss (convex/non-convex) and any model (linear/non-linear, parametric/non-parametric) for complementary learning.

Next, we show the relationship between our proposed framework and previous complementary-label learning \citep{ishida17nips}.
\begin{corollary}\label{corollary}
If one-versus-all loss \eqref{comp-ova} or pairwise comparison loss \eqref{comp-pc} is used with binary loss function that satisfy $s(z)+s(-z)=1$, the classification risk can be written as,
\begin{align}
\overline{R}(\bmg;\overline{\ell})=(K-1)\mathbb{E}_{\overline{\mathcal{D}}}\big[\overline{\ell}\big(\overline{Y},\bmg(X)\big)\big]-M_1+M_2,
\end{align}
where $M_1$ and $M_2$ are non-negative constants that satisfy $\sum^K_{\overline{y}=1}\overline{\ell}\big(\overline{y},\bmg(x)\big)=M_1$ for all $x$ and $\overline{\ell}\big(\overline{y},\bmg(x)\big)+\ell\big(\overline{y},\bmg(x)\big)=M_2$ for all $x$ and $\overline{y}$.
\end{corollary}
Proof can be found in Appendix \ref{sec:corollary2proof}.  Since this is equivalent to the first two Theorems in \citet{ishida17nips}, our proposed version is a generalization of the previous unbiased complementary-label learning framework.

The key idea of the proof in Theorem \ref{theorem:R(f)} is to not rely on the condition
that $\sum_{k=1}^K\overline{\ell}\big(k,\bmg(x)\big)$ is a constant for all $x$,
used in \citet{ishida17nips}, which is inspired by the property of binary 0-1 loss $s_{0-1}$,
where $s_{0-1}(z)$ is $1$ if $z<0$ and $0$ otherwise.
Such a technique was also used when designing unbiased risk estimators for learning from positive and unlabeled data in a binary classification setup \citep{christo14nips}, but was later shown to be unnecessary \citep{christo15icml}.
Note that Theorem \ref{theorem:R(f)} can be regarded as a special case of a framework proposed for learning from weak labels \citep{jesus14ecml}.

By using \eqref{scalar_complementary_loss}, the classification risk can be written as
\begin{align}
\label{risk:general_complementary}
R(\bmg; \ell)
  = \sum^K_{k=1} \overline{\pi}_k \mathbb{E}_{\overline{P}_k} \Big[& -(K-1)\cdot \ell\big(k,\bmg(X)\big)\nonumber\\
  & + \sum^K_{j=1}\ell\big(j,\bmg(X)\big)\Big].
\end{align}
Here, we rearrange our complementarily labeled dataset as $\{\mathcal{X}_k\}_{k=1}^{K}$, where $\mathcal{X}_k$ denotes the samples complementarily labeled as class $k$.
Then, this expression of the classification risk can be approximated by,
\begin{align}
\label{emp:general}
\widehat{R}(\bmg;\ell)
= \sum^K_{k=1} \frac{\widehat{\overline{\pi}}_k}{\left|\mathcal{X}_k\right|}\sum_{x_i\in\mathcal{X}_k} \Big[ &-(K-1)\cdot \ell\big(k,\bmg(x_i)\big) \nonumber\\
&+ \sum^K_{j=1}\ell\big(j,\bmg\big(x_i)\big)\Big],
\end{align}
where $n_k$ is the number of patterns complementarily labeled as the $k$th class.

\subsection{Necessity of risk correction}
\label{sec:necessity}
The original expression of the classification risk \eqref{risk:ordinary} includes an expectation over non-negative loss $\ell:[K]\times \mathbb{R}^K\rightarrow \mathbb{R}_+$, so the risk and its empirical approximator are both lower-bounded by zero.
On the other hand, the expression \eqref{risk:general_complementary} derived above contains a negative element. Although \eqref{risk:general_complementary} is still non-negative by definition, due to the negative term, its empirical estimator can go negative, leading to over-fitting.

\begin{figure*}
\centering
  \centering
  \includegraphics[bb = 0 0 1376 259, scale = 0.50]{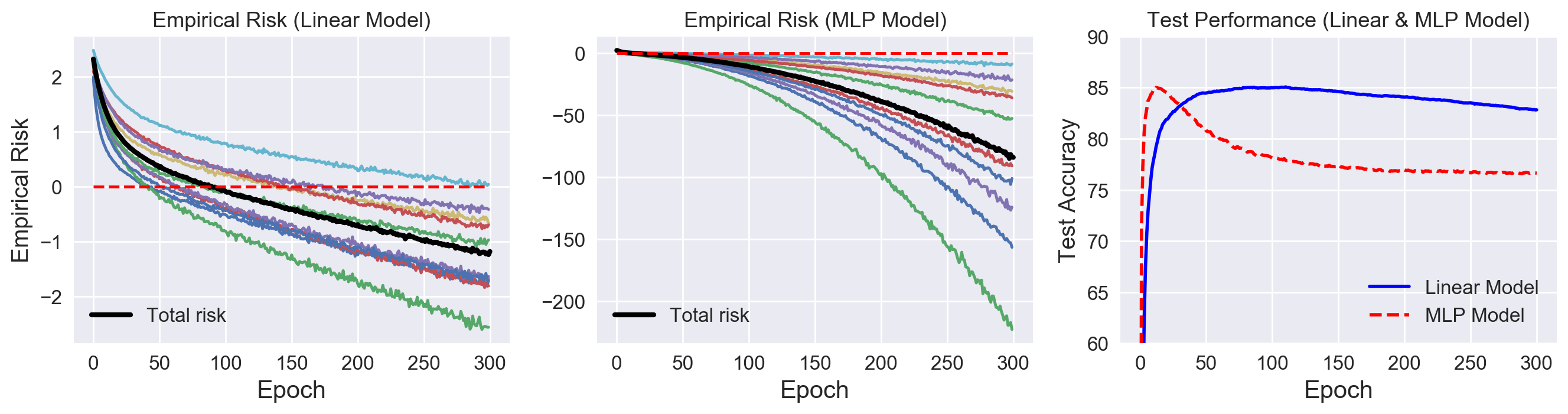}
  \caption{The left and middle graphs shows the total risk \eqref{emp:general} (in black color) and the risk decomposed into each \emph{ordinary} class term \eqref{eq:counterparts} (in other colors) for training data with linear and MLP models, respectively.
  The right graph shows the corresponding test accuracy for both models.}
  \label{fig:negativerisk}
\end{figure*}

We elaborate on this issue with an illustrative numerical example.
In the left graph of Figure \ref{fig:negativerisk}, we show an example of training a linear model trained on the handwritten digits dataset MNIST\footnote{See \url{http://yann.lecun.com/exdb/mnist/}.}, with complementary labels generated to satisfy \eqref{unbiased-complementary-label-assumption}.  We used \emph{Adam} \citep{kingma15iclr} for optimization with learning rate $5\mathrm{e}-5$, mini-batch size of 100, and weight decay of $1\mathrm{e}-4$ with 300 epochs.  The empirical classification risk \eqref{emp:general} is shown in black.
We can see that the empirical classification risk continues decreasing and can go below zero at around 100 epochs.  The test accuracy on the right graph hits the peak also at around epoch 100 and then the accuracy gradually deteriorates.

This issue stands out even more significantly when we use a flexible model.  The middle graph shows the empirical classification risk for a multilayer perceptron (MLP) with one hidden layer ($500$ units), where \emph{ReLU} \citep{relu} was used as the activation function.
The optimization setup was the same as the case of the linear model above.
We can see the empirical risk decreasing much more quickly and going negative.
Correspondingly, as the right graph shows, the test accuracy drops significantly after the empirical risk goes negative.

In fact, a similar issue is already implicit in the original paper by \citet{ishida17nips}: According to Corollary \ref{corollary} (or Theorem 1 in \citet{ishida17nips}), the unbiased risk estimator includes subtraction of a positive constant term which increases with respect to the number of classes.
This means that the learning objective of \citet{ishida17nips} has a (negative) lower bound.

\subsection{Non-negative risk estimator}
\label{sc:nn_risk_estimator}
As we saw in Section \ref{sec:necessity}, our risk estimator can suffer from overfitting due to the non-negative issue.
Here, we propose a correction to the risk estimator to overcome this problem.

Each term in the risk with ordinary labels (right-hand side of \eqref{risk:class_prior_decomposition}), which corresponds to each class, is non-negative.
We can reformulate \eqref{risk:general_complementary} in order to show the counterpart for each non-negative term in the right-hand side of \eqref{risk:class_prior_decomposition} for complementarily labeled data as
\begin{align}
  R(\bmg;\ell)
  \label{eq:counterparts}
  = \sum_{k=1}^K \Big[ -(K-1) \overline{\pi}_k\cdot \mathbb{E}_{\overline{P}_k}\big[\ell\big(k,\bmg(X)\big)\big]\nonumber\\
  +\sum^K_{j=1} \overline{\pi}_j \cdot \mathbb{E}_{\overline{P}_j}\big[\ell\big(k,\bmg(X)\big)\big] \Big].
\end{align}
These counterparts \eqref{eq:counterparts} were originally non-negative when ordinary labels were used.
In the left and middle graphs of Figure \ref{fig:negativerisk}, we plot the decomposed risks with respect to each \emph{ordinary} class \eqref{eq:counterparts} (shown in different colors).
We can see that the decomposed risks for all classes become negative eventually.
Based on this observation, our basic idea for correction is to enforce non-negativity for each ordinary class, with the expression based on complementary labels.
More specifically, we propose a non-negative version by
\begin{align}
\label{non-negative-risk}
  \sum^K_{k=1} \max\Big\{ 0, \Big[-(K-1) \overline{\pi}_k \cdot \mathbb{E}_{\overline{P}_k}\big[\ell\big(k,\bmg(X)\big)\big]\nonumber\\
  +\sum^K_{j=1} \overline{\pi}_j \cdot \mathbb{E}_{\overline{P}_j}\big[\ell\big(k,\bmg(X)\big)\big] \Big]\Big\}.
\end{align}
\eqref{non-negative-risk} is equivalent to \eqref{eq:counterparts}, since $\max\{0,a\}=a$ if $a$ is non-negative.
By using the datasets used for \eqref{emp:general}, this non-negative risk can be na\"ively approximated by the sample average as
\begin{align}
\label{emp_nn}
  \sum^K_{k=1} \max \Big\{ 0, \Big[-(K-1)\cdot\frac{\overline{\pi}_k}{\left|\mathcal{X}_k\right|}\sum_{x_i\in\mathcal{X}_k} \ell(k,\bmg(x_i))\nonumber\\
  + \sum^K_{j=1} \frac{\overline{\pi}_j}{\left|\mathcal{X}_j\right|} \sum_{x_{i'}\in\mathcal{X}_j}\ell(k,\bmg(x_{i'})) \Big]\Big\}.
\end{align}
The empirical version of \eqref{eq:counterparts} may suffer from a negative objective, but \eqref{emp_nn} is non-negative (even though their population versions are equivalent.)

Enforcing the reformulated risk to become non-negative was previously explored in \citet{kiryo17nips}, in the context of binary classification from positive and unlabeled data.  The positive class risk is already bounded below by zero in their case (because they have true positive labels), so there was a max operator only on the negative class risk.  We follow their footsteps, but since our setting is a multi-class scenario and also differs by not having \emph{any} true labels, we put a max operator on each of the $K$ classes.
\begin{algorithm}[t]
  \caption{Complementary-label learning with gradient ascent}
  \label{alg:ascent}
  \begin{algorithmic}
    \STATE \textbf{Input:} complementarily labeled training data $\{\mathcal{X}_k\}_{k=1}^{K}$, where $\mathcal{X}_k$ denotes the samples complementarily labeled as class $k$;
    \STATE \textbf{Output:} model parameter $\theta$ for $g(x;\theta)$
  \end{algorithmic}
  \begin{algorithmic}[1]
    \STATE Let $\mathcal{A}$ be an external SGD-like stochastic optimization algorithm such as \citet{kingma15iclr}
    \STATE \textbf{while} no stopping criterion has been met:
    \STATE \quad Shuffle $\{\mathcal{X}_j\}^K_j$ into $B$ mini-batches;
    \STATE \quad \textbf{for} $b=1$ \textbf{to} $B$:
    \STATE \qquad Denote $\{\mathcal{X}_j^b\}$ as the $b$-th mini-batch for complementary class $j$
    \STATE \qquad Denote $r^b_k(\theta) = -(K-1)\overline{\pi}_k\cdot\widehat{\mathbb{E}}_{\overline{P}_k}[\ell(k,\bmg);\mathcal{X}^b_k] + \sum^K_{j=1}\overline{\pi}_j\cdot\widehat{\mathbb{E}}_{\overline{P}_j}[\ell(k,\bmg);\mathcal{X}^b_j]$
    \STATE \qquad \textbf{if} $\min_k[r^b_1(\theta),\ldots,r^b_k(\theta),\ldots,r^b_K(\theta)]>-\beta$:
    \STATE \qquad\quad Denote $L^b(\theta) = \sum^K_{k=1} r^b_k(\theta)$
    \STATE \qquad\quad Set gradient $\nabla_\theta L^b(\theta)$;
    \STATE \qquad\quad Update $\theta$ by $\mathcal{A}$ with its current step size $\eta$;
    \STATE \qquad \textbf{else}:
    \STATE \qquad\quad Denote $\widetilde{L}^b(\theta) = \sum^K_{k=1}\min\{-\beta, r^b_k(\theta)\}$
    \STATE \qquad\quad Set gradient $- \nabla_\theta \widetilde{L}^b(\theta)$;
    \STATE \qquad\quad Update $\theta$ by $\mathcal{A}$ with a discounted step size $\gamma\eta$;
  \end{algorithmic}
\end{algorithm}
\subsection{Approximate non-negative risk estimator}
\label{sc:implementation}
\begin{figure*}[t]
\vspace{-5mm}
\centering
\subfigure[MNIST, linear]{\includegraphics[bb = 0 0 351 348, scale = 0.3339]{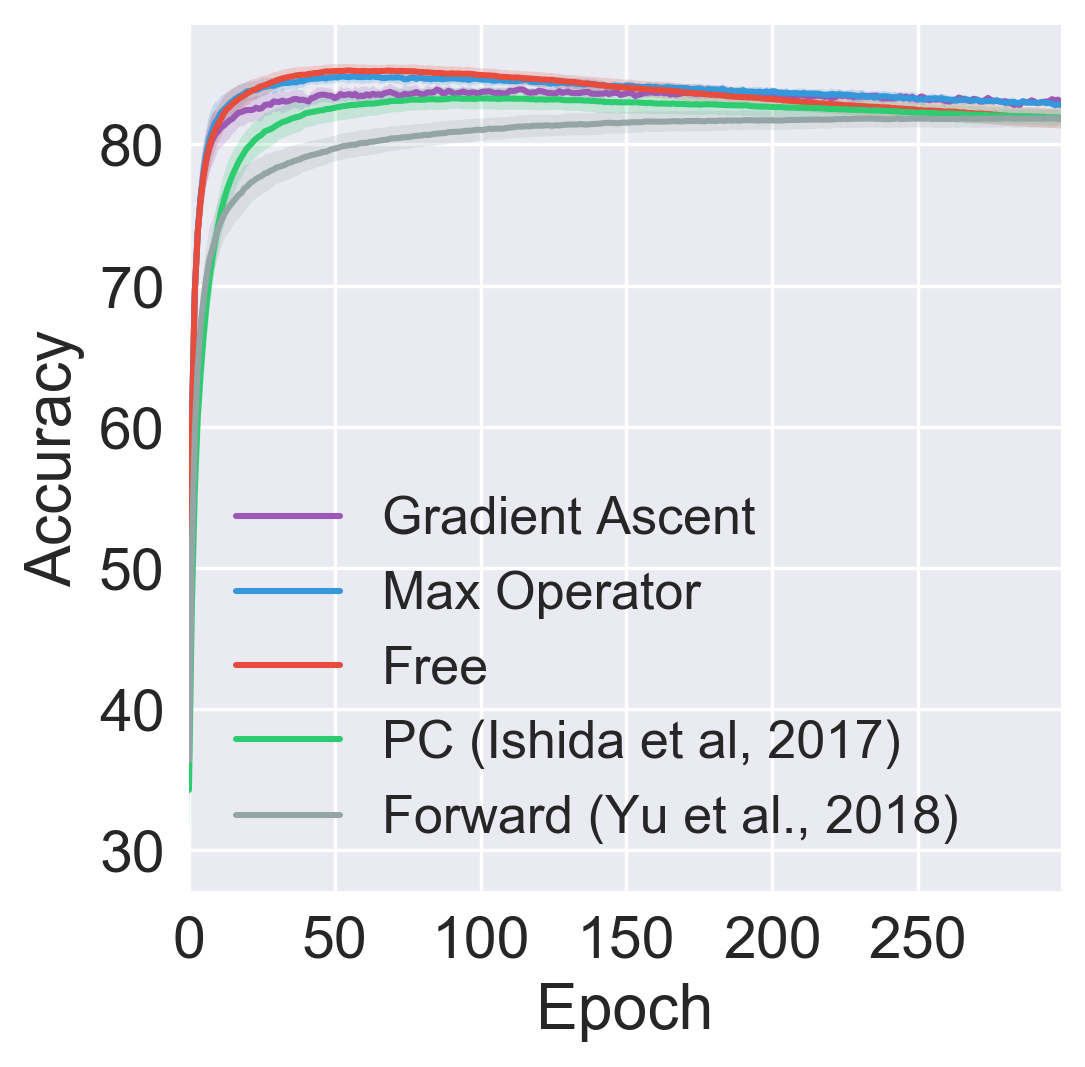}}
\hspace{0.001\textwidth}
\subfigure[MNIST, MLP]{\includegraphics[bb = 0 0 351 348, scale = 0.3339]{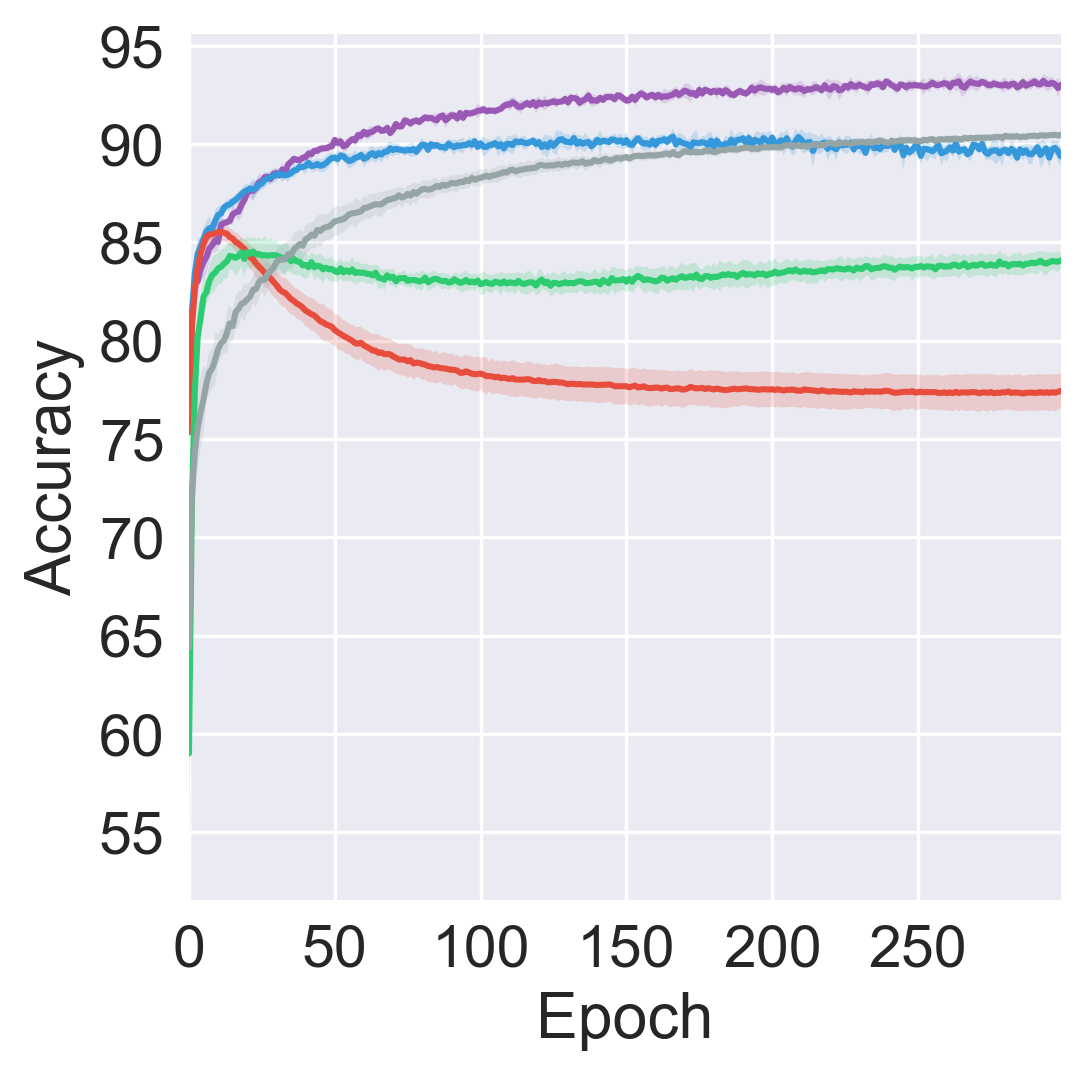}}
\hspace{0.001\textwidth}
\subfigure[Fashion-MNIST, linear]{\includegraphics[bb = 0 0 351 348, scale = 0.3339]{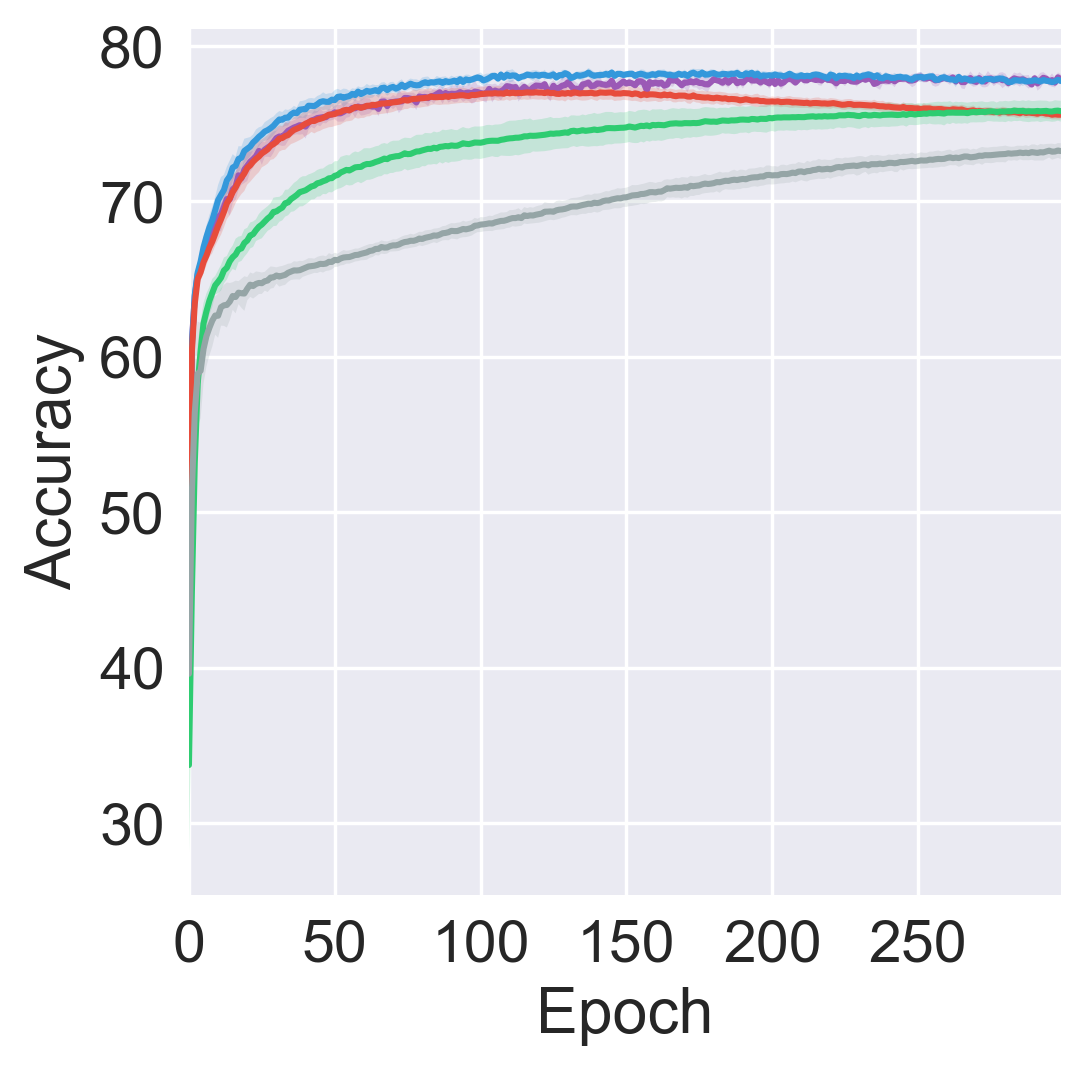}}
\hspace{0.001\textwidth}
\subfigure[Fashion-MNIST, MLP]{\includegraphics[bb = 0 0 351 348, scale = 0.3339]{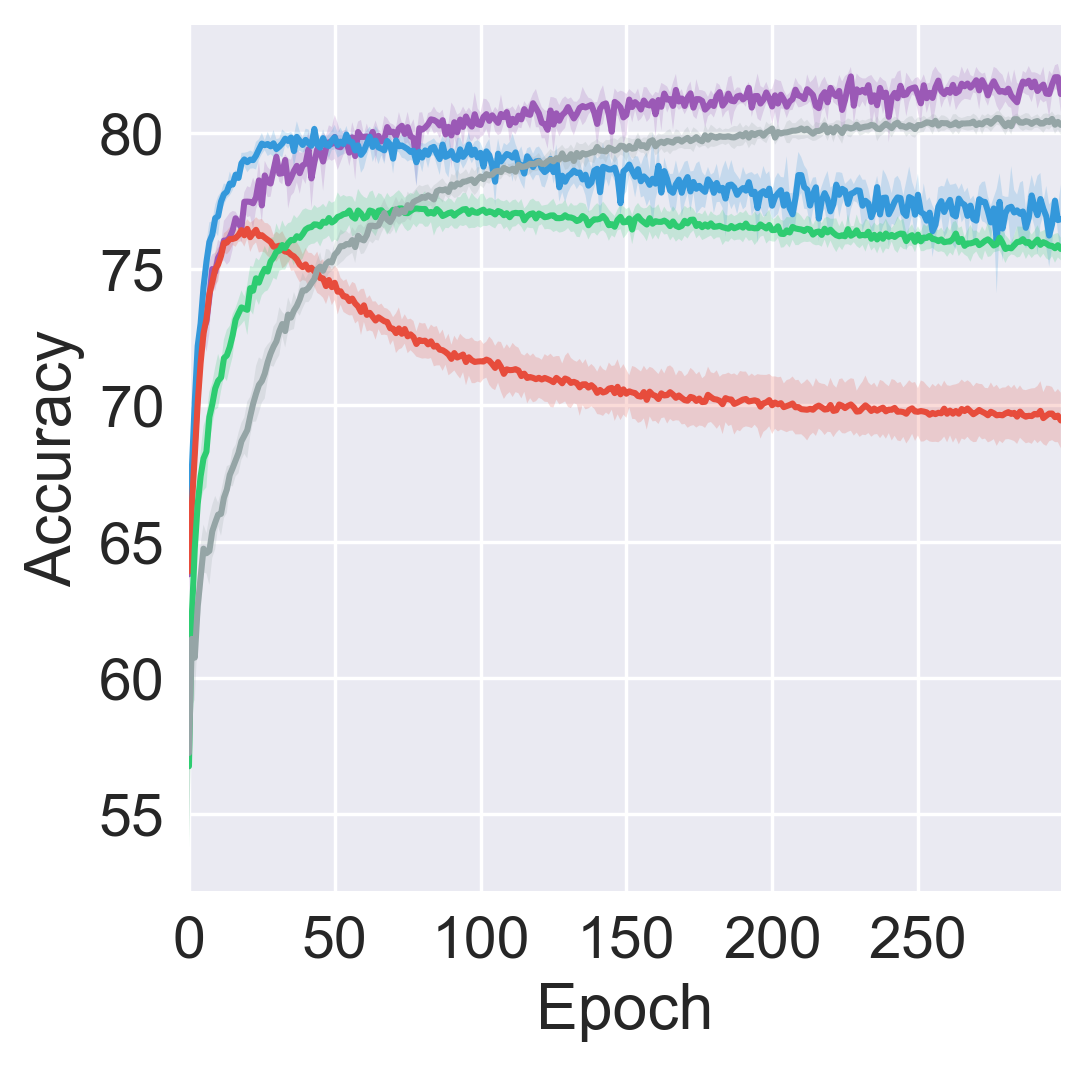}}\\
\vspace{-3mm}
\subfigure[Kuzushiji-MNIST, linear]{\includegraphics[bb = 0 0 351 348, scale = 0.3339]{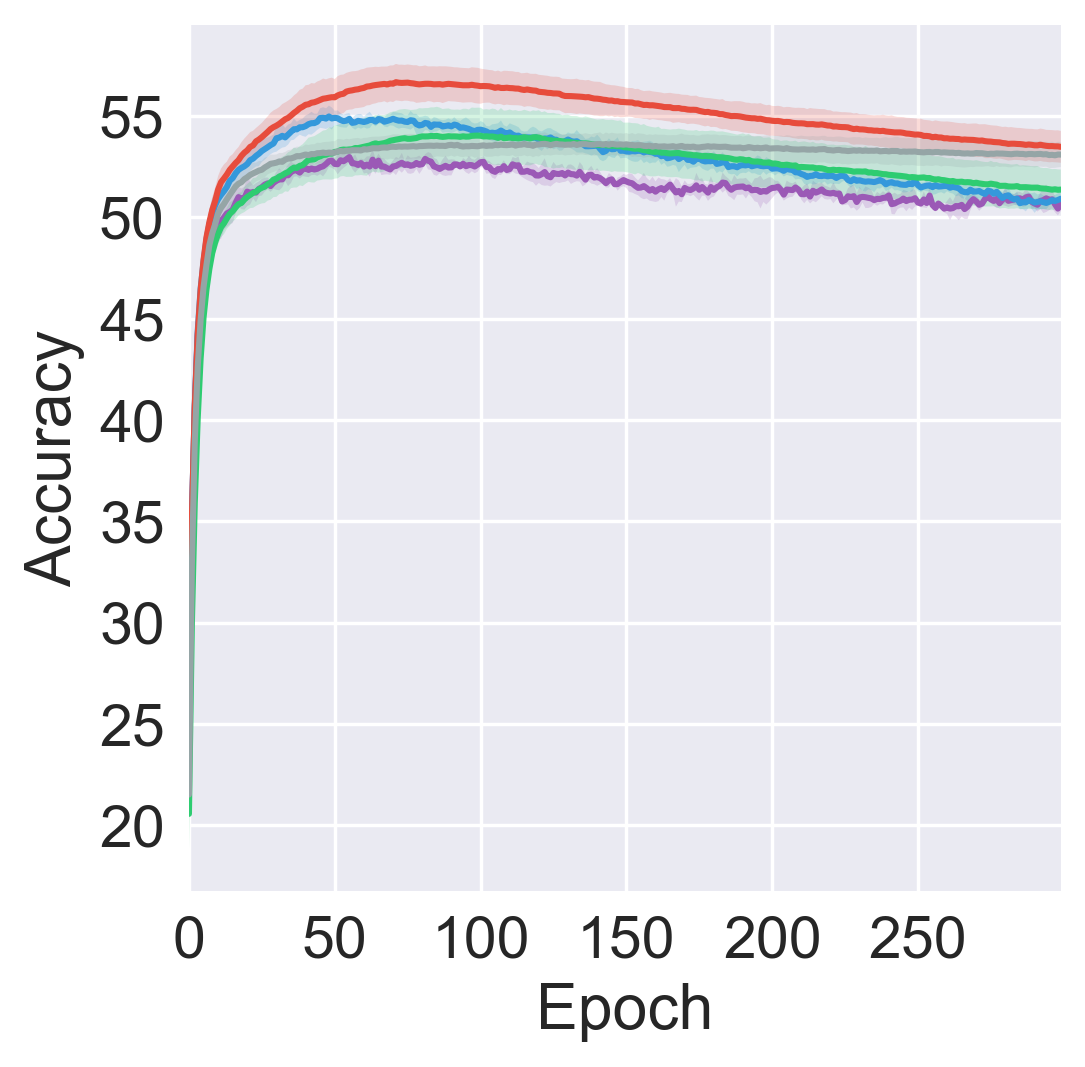}}
\hspace{0.001\textwidth}
\subfigure[Kuzushiji-MNIST, MLP]{\includegraphics[bb = 0 0 351 348, scale = 0.3339]{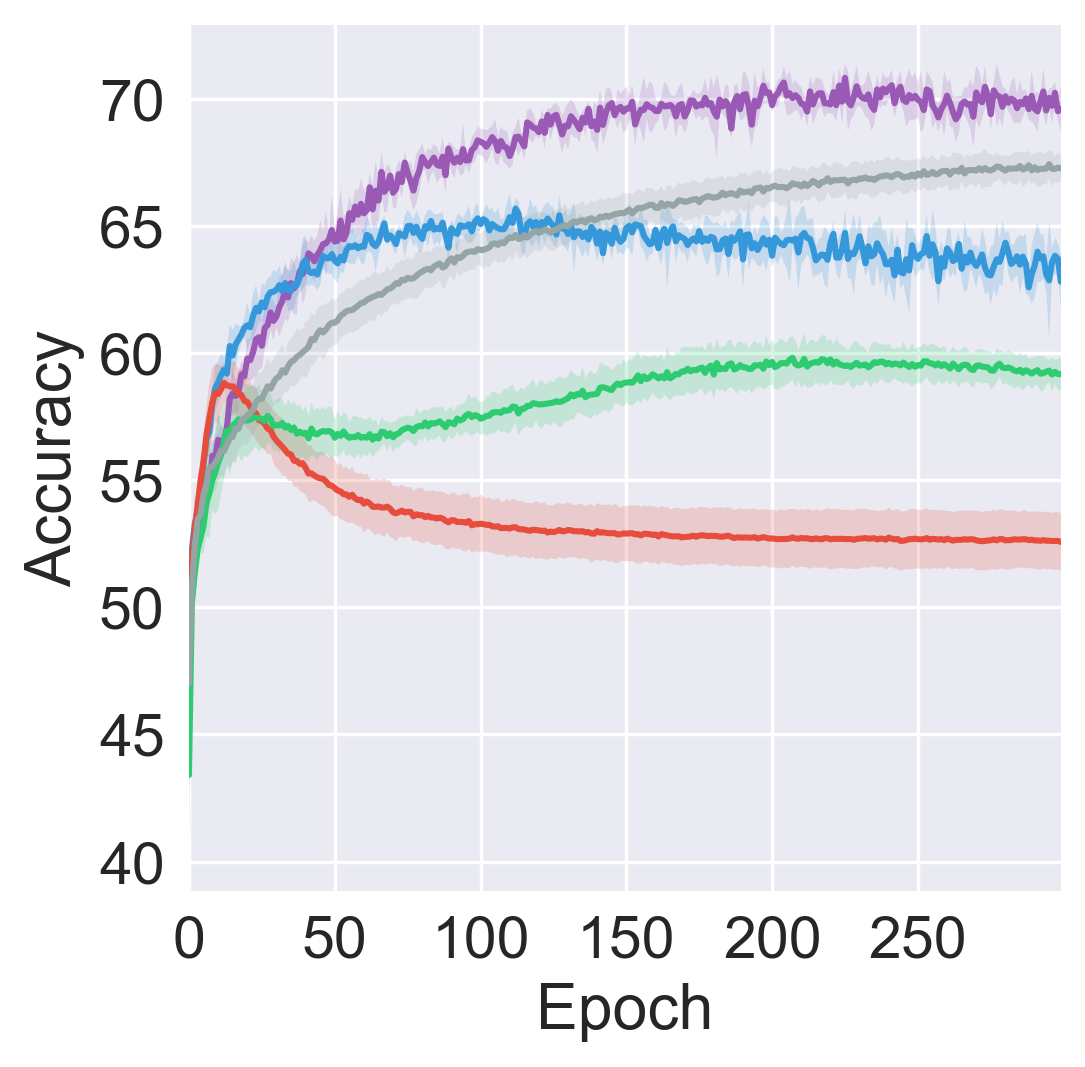}}
\hspace{0.001\textwidth}
\subfigure[CIFAR-10, DenseNet]{\includegraphics[bb = 0 0 351 348, scale = 0.3339]{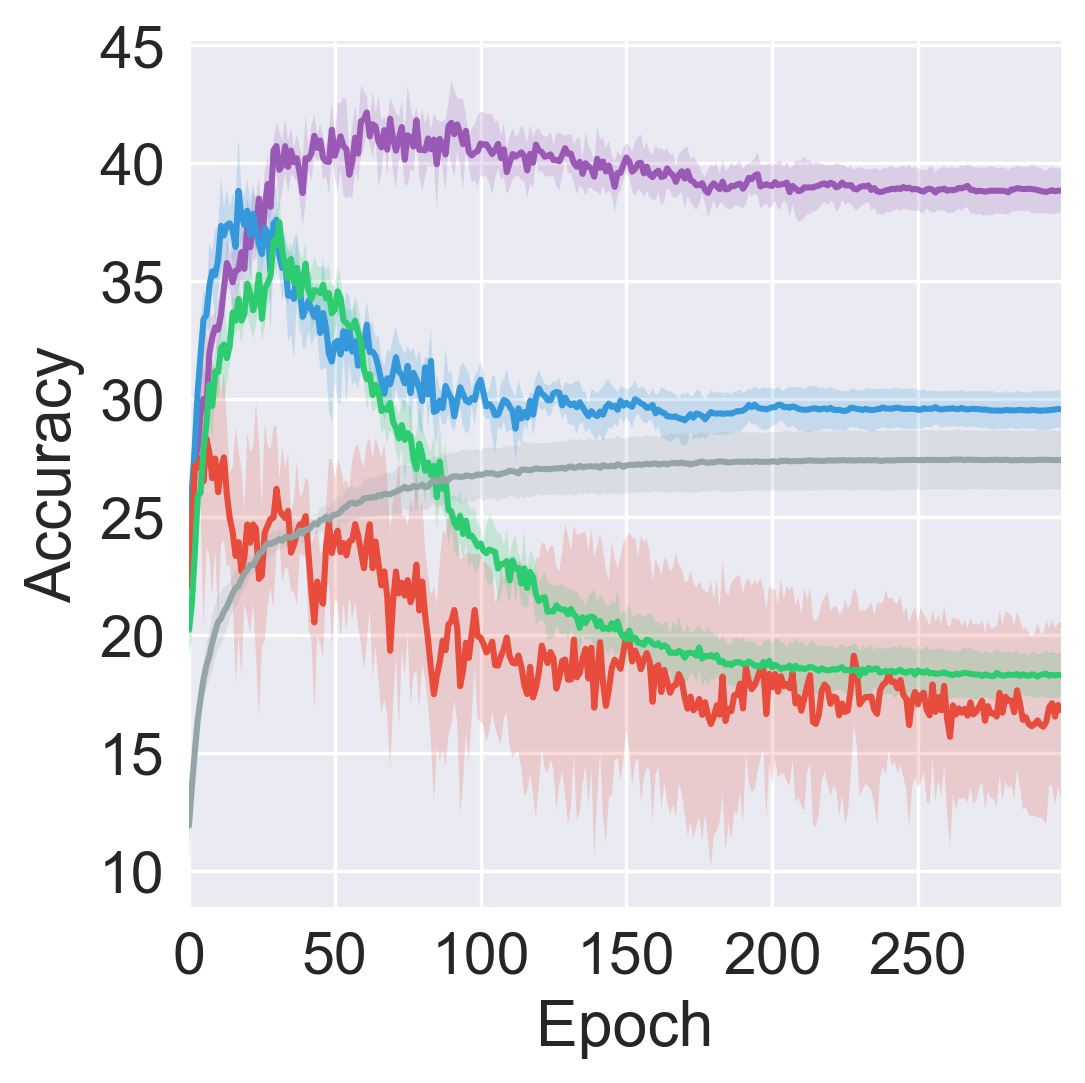}}
\hspace{0.001\textwidth}
\subfigure[CIFAR-10, ResNet]{\includegraphics[bb = 0 0 351 348, scale = 0.3339]{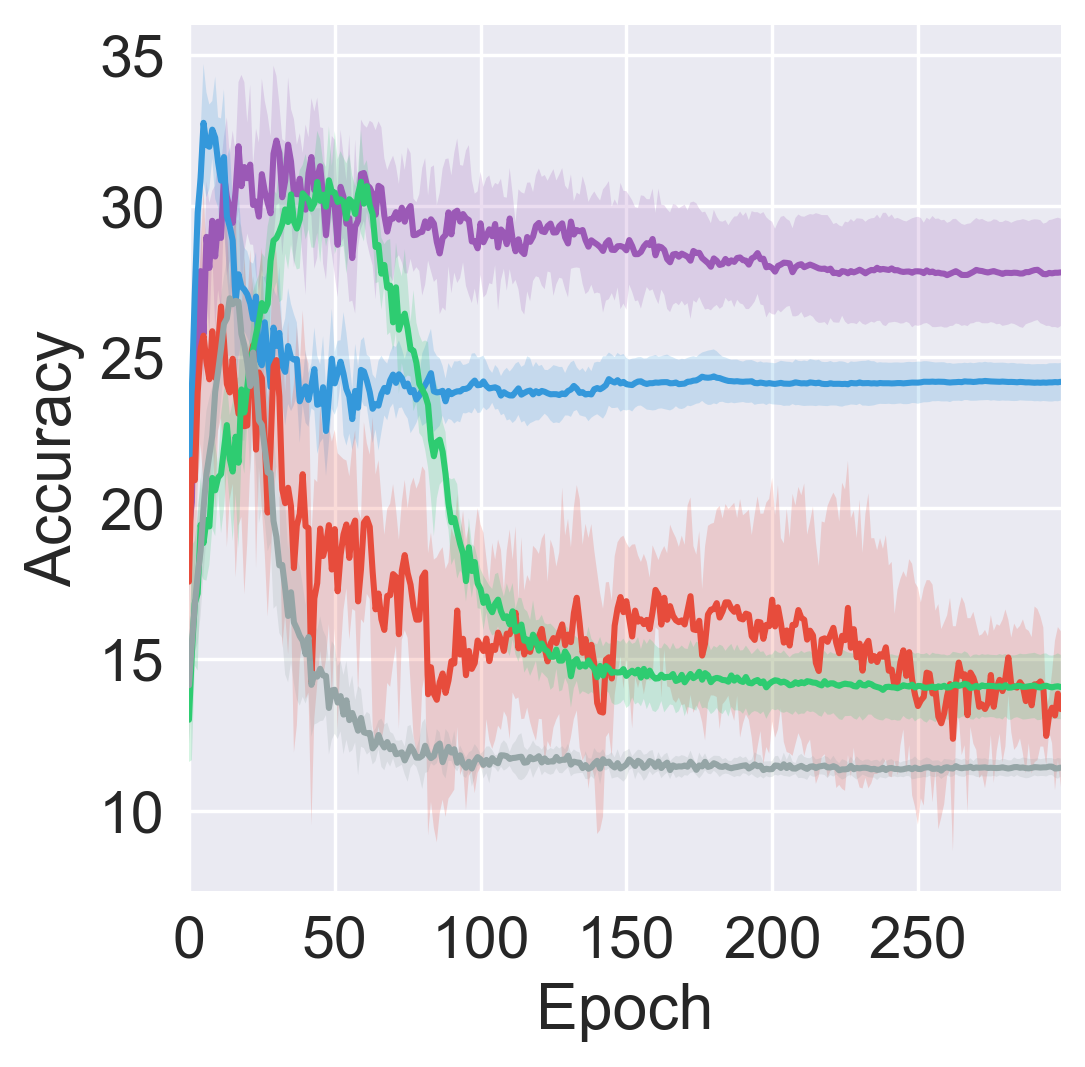}}
\caption{Experimental results for various datasets and models.  Dark colors show the mean accuracy of 5 trials and light colors show standard deviation.}\label{fig:results}
\end{figure*}
\paragraph{Implementation with max operator}
We now illustrate how to design a practical implementation under stochastic optimization for our non-negative risk estimator.
An unfortunate issue is that the minimization of \eqref{emp_nn} is not point-wise due to the max-operator, thus cannot be used directly for stochastic optimization methods with mini-batch.  However, an upper bound of the risk can be minimized in parallel by using mini-batch as the following,
\begin{align}
\label{emp_nn_minibatch}
\frac{1}{B}\sum_{b=1}^N\sum_{k=1}^K \max \Big\{0,
-(K-1)\overline{\pi}_k\cdot\widehat{\mathbb{E}}_{\overline{P}_k}\big[\ell\big(k,\bmg(X)\big);\mathcal{X}^b_k\big]\nonumber\\
+ \sum^K_{j=1}\overline{\pi}_j\cdot\widehat{\mathbb{E}}_{\overline{P}_j}\big[\ell\big(k,\bmg(X)\big);\mathcal{X}^b_j\big]
\Big\},
\end{align}
where $\widehat{\mathbb{E}}$ is the empirical version of the expectation and $B$ is the number of mini-batches.
\paragraph{Implementation with gradient ascent}
If the objective is negative for a certain mini-batch, the previous implementation based on the max operator will prevent the objective to further \emph{decrease}.  However, if the objective is already negative, that mini-batch has already started to overfit.  The max operator cannot contribute to decrease the degree of overfitting. From this perspective, there is still room to improve the overfitting issue, and it would be preferable to \emph{increase} itself to make this mini-batch less overfitted.

Our idea is the following.
We denote the risk that corresponds to the $k$th ordinary class for the $i$th mini-batch as
\begin{align*}
r^b_k(\theta) = -(K-1)\overline{\pi}_k\cdot\widehat{\mathbb{E}}_{\overline{P}_k}[\ell\big(k,\bmg(X)\big);\mathcal{X}^b_k]\nonumber\\
+ \sum^K_{j=1}\overline{\pi}_j\cdot\widehat{\mathbb{E}}_{\overline{P}_j}\big[\ell\big(k,\bmg(X)\big);\mathcal{X}^b_j\big],
\end{align*}
and the total risk as
$L^b(\theta) = \sum^K_{k=1}r^b_k(\theta)$.
When $\min_k\{r^b_k(\theta)\}_{k=1}^K\geq-\beta$, we conduct gradient descent as usual with gradient $\nabla_\theta L^b(\theta)$.
On the other hand, if $\min_k\{r^b_k(\theta)\}_{k=1}^K<-\beta$, we first squash the class-decomposed risks over $-\beta$ to $-\beta$ with a min operator, and then sum the results:
\begin{align*}
\widetilde{L}^b(\theta) = \sum^K_{k=1}\min\{-\beta, r^b_k(\theta)\}.
\end{align*}
Next we set the gradient in the opposite direction with $-\nabla_\theta \widetilde{L}^b(\theta)$.  Conceptually, we are going \emph{up} the gradient $\nabla_\theta \widetilde{L}^b(\theta)$ for \emph{only} the class-decomposed risks below $-\beta$, to avoid the class-decomposed risks that are already large to further increase.
Note that $\beta$ is a hyper-parameter that controls the tolerance of negativity.  $\beta=0$ would mean there is zero tolerance, but in practice we can also have $-\beta\neq0$ for a threshold that allows some negative ($-\beta<0$) or positive ($-\beta>0$) amount.
The procedure is shown in detail in Algorithm \ref{alg:ascent}.
\begin{table*}[t]
  \center
  \small
  \caption{Test mean and standard deviation of the classification accuracy for 4 trials.  Method name outside (inside) parenthesis shows the criterion of training (validation) objective.  Best is shown in {\bf bold} or \underline{underline} for column 2$\sim$4 or column 2$\sim$6, respectively.}
  \label{tb:validation}
  \tabcolsep=0.2cm
    \begin{tabular}{l|ccc|cc}\toprule
    Dataset & \emph{GA (Free)} & \emph{PC (PC)} & \emph{Fwd (Fwd)} & \emph{PC (Free)} & \emph{Fwd (Free)} \\\midrule
    MNIST&$88.1\pm2.5\%$&$79.3\pm3.3\%$&\bm{$88.7\pm0.3\%$}&$80.2\pm2.9\%$&\underline{$89.4\pm0.4\%$}\\\midrule
    Fashion&$\bm{\underline{78.7\pm1.4\%}}$&$74.7\pm1.6\%$&$77.5\pm1.2\%$&$75.7\pm1.2\%$&$73.5\pm5.5\%$ \\\midrule
    Kuzushiji&\bm{$63.8\pm1.1\%$}&$56.7\pm4.9\%$&$62.0\pm1.1\%$&$56.1\pm4.2\%$&\underline{$65.4\pm1.7\%$} \\\midrule
    CIFAR-10&$\bm{\underline{36.8\pm0.6\%}}$&$33.4\pm2.0\%$&$30.8\pm1.6\%$&$25.9\pm7.6\%$&$30.8\pm1.7\%$ \\\bottomrule
  \end{tabular}
\end{table*}
\section{Experiments}
\label{sc:experiments}
In this section, we compare the 3 methods that we have proposed in Section \ref{sc:proposed_method}, which are \emph{Free} (Unbiased risk estimator that is loss assumption free, based on Eq. \eqref{emp:general}), \emph{Max Operator} (based on Eq. \eqref{emp_nn_minibatch}), and \emph{Gradient Ascent} (based on Alg.\ref{alg:ascent}).
For \emph{Gradient Ascent}, we used $\beta=0$ and $\gamma=1$ for simplicity.
Mini-batch size was set to $256$.
We also compare with two baseline methods: Pairwise comparison (\emph{PC}) with ramp loss from \citet{ishida17nips} and \emph{Forward} correction from \citet{yu17eccv}.
For training, we used only complementarily labeled data, which was generated so that the assumption of \eqref{unbiased-complementary-label-assumption} is satisfied.
This is straightforward when the dataset has a uniform (ordinarily-labeled) class prior, because it reduces to just choosing a class randomly other than the true class.

In Appendix \ref{sec:datasets}, we explain the details of the datasets used in the experiments: MNIST, Fashion-MNIST, Kuzushiji-MNIST, and CIFAR-10.
The implementation is based on Pytorch\footnote{\url{https://pytorch.org}} and our demo code is available online\footnote{\url{https://github.com/takashiishida/comp}}.

\subsection{Comparison of all epochs during training}
\label{sec:all_epochs_exp}
\paragraph{Setup}
For MNIST, Fashion-MNIST, and Kuzushiji-MNIST, a linear-in-input model with a bias term and a MLP model ($d-500-1$) was trained with softmax cross-entropy loss function (except \emph{PC}) for $300$ epochs.
Weight decay of $1\mathrm{e}-4$ for weight parameters and learning rate of $5\mathrm{e}-5$ for Adam \citep{kingma15iclr} was used.

For CIFAR-10, DenseNet \citep{huang2017cvpr} and ResNet-34 \citep{he2016cvpr} were used with weight decay of $5\mathrm{e}-4$ and initial learning rate of $1\mathrm{e}-2$.  For optimization, stochastic gradient descent was used with the momentum set to $0.9$.  Learning rate was halved every $30$ epochs.

\paragraph{Results}
We show the accuracy for all $300$ epochs on test data to demonstrate how the issues discussed in Section \ref{sec:necessity} appear and how different implementations in Section \ref{sc:implementation} are effective. In Figure \ref{fig:results}, we show the mean and standard deviation of test accuracy for 4 trials on test data evaluated with ordinary labels.

First we compare our 3 proposed methods with each other.
For linear models in MNIST, Fashion-MNIST, and Kuzushiji-MNIST, all proposed methods work similarly.
However in the case of using a more flexible MLP model or using DenseNet/ResNet in CIFAR-10, we can see that \emph{Free} is the worst, \emph{Max Operator} is better and \emph{Gradient Ascent} is the best out of the proposed three methods for most of the epochs (\emph{Free} $<$ \emph{Max Operator} $<$ \emph{Gradient Ascent}).
These results are consistent with the discussions of overfitting in Section \ref{sec:necessity} and the motivations for different implementations in Section \ref{sc:implementation}.

Next, we compare with baseline methods.
For linear models, all methods have similar performance.
However for deep models (MLP, DenseNet, and ResNet), the superiority stands out for \emph{Gradient Ascent} for all datasets.

\subsection{Experiments with validation process}
\label{sec:validation_experiments}
\paragraph{Setup}
Next, we perform experiments with a train, validation, and test split.  The dataset is constructed by splitting the original training data used in the previous experiments into train/validation with a 9:1 ratio.  Note that the validation data only has complementary labels since it is splitted from the set of complementarily labeled training data.
We use the same MLP models for MNIST, Fashion-MNIST, and Kuzushiji-MNIST.
We use DenseNet for CIFAR-10.

Since \emph{Gradient Ascent (GA)} seemed to work better than \emph{Free} and \emph{Max Operator} previously, we omit \emph{Free} and \emph{Max Operator} and compare \emph{GA} with baseline methods (\emph{PC} and \emph{Forward(Fwd)}).
For the validation objective, we used the corresponding criterion for each method, which is shown in the first 3 columns with parenthesis, in Table \ref{tb:validation}.
We also conducted experiments using our proposed general unbiased estimator \emph{Free} as the validation criterion for baseline methods (\emph{PC} and \emph{Fwd}), which is shown in the last 2 columns in Table \ref{tb:validation}.
SGD with momentum of $0.9$ was used for $250$ epochs.
Weight-decay was fixed to $1e-4$ and learning rate candidates are \{$1\mathrm{e}-4$, $5\mathrm{e}-4$, $1\mathrm{e}-3$, $5\mathrm{e}-3$, $1\mathrm{e}-2$, $5\mathrm{e}-2$\} for CIFAR-10 and \{$5\mathrm{e}-5$, $1\mathrm{e}-4$, $5\mathrm{e}-4$, $1\mathrm{e}-3$, $5\mathrm{e}-3$, $1\mathrm{e}-2$\} for other datasets.
For CIFAR-10, we added learning rate decay with the same settings from Section \ref{sec:all_epochs_exp}.

\paragraph{Results}
In Table \ref{tb:validation}, we showed the mean and standard deviation of test accuracy for 4 trials, with the model that gave the best validation score out of all epochs for all hyper-parameter candidates.
By comparing the first 3 columns, \emph{GA} seems to work well.
We can also observe that in most cases, \emph{PC (Free)} and \emph{Fwd (Free)} performs similarly or better than \emph{PC (PC)} and \emph{Fwd (Fwd)}, respectively. This confirms the discussion in earlier sections that our general unbiased risk estimator is useful not only as a learning objective, but also useful as a validation objective for baseline methods.

\section{Conclusion}\label{sc:conclusion}
We first proposed a general risk estimator for learning from complementary labels that does not require restrictions on the form of the loss function or the model.
However, since the proposed method suffers from overfitting, we proposed a modified version to alleviate this issue in two ways and have better performance.
At last, we conducted experiments to show our proposed method outperforms or is comparable to current state-of-the-art methods for various benchmark datasets and for both linear and deep models.

Recently, \emph{complementary-label learning} has been applied to \emph{online learning} \citep{kaneko19arxiv}, \emph{generative discriminative learning} \citep{xu19arxiv}, and \emph{medical image segmentation} \citep{rezaei19mta}.  This implies applying the idea of complementary labels to other domains may be useful, which can be an interesting future direction.
\newpage
\section*{Acknowledgments}
TI was supported by Sumitomo Mitsui DS Asset Management. MS was supported by JST CREST JPMJCR1403.
We thank the anonymous reviewers for the helpful suggestions.

\bibliography{example_paper}
\bibliographystyle{icml2019}

\clearpage
\appendix

\section{Proof of Theorem \ref{theorem:R(f)}}\label{sec:theorem1proof}
\begin{proof}
First of all,
\begin{align*}
&\mathbb{P}(X, \overline{Y}=\overline{y})
=\frac{1}{K-1}\sum_{y\neq\overline{y}}\mathbb{P}(X, Y=y)\\
&=\frac{1}{K-1}\Big(\sum^K_{y=1}\mathbb{P}(X,Y=y)-\mathbb{P}(X,Y=\overline{y})\Big)\\
&=\frac{1}{K-1}\big(\mathbb{P}(X)-\mathbb{P}(X,Y=\bar{y})\big).
\end{align*}
The first equality holds since the marginal distribution is equivalent for $\mathcal{D}$ and $\overline{\mathcal{D}}$ and we assume \eqref{unbiased-complementary-label-assumption}.
Consequently,
\begin{align*}
&\mathbb{P}(\overline{Y}=\overline{y}|X=x)
= \frac{\mathbb{P}(X=x, \overline{Y}=\overline{y})}{\mathbb{P}(X=x)}\\
&=\frac{1}{K-1}\cdot \Big(1- \frac{\mathbb{P}(X,Y=\overline{y})}{\mathbb{P}(X=x)}\Big)\\
&=\frac{1}{K-1}\cdot \big(1-\mathbb{P}(Y=\overline{y}|X=x)\big)\\
&=-\frac{1}{K-1}\mathbb{P}(Y=\overline{y}|X=x)+\frac{1}{K-1}.
\end{align*}
More simply, we have
$\bm{\eta}(x) = -(K-1) \overline{\bm{\eta}}(x) + \bm{1}$.
Finally, we transform the classification risk,
\begin{align*}
R(g;\ell)
&=\mathbb{E}_{(X,Y)\sim \mathcal{D}}[\ell(Y,\bmg(X))]
=\mathbb{E}_{X\sim M}[\bm{\eta}^\top \bm{\ell}(\bmg(X))]\\
&=\mathbb{E}_{X\sim M}\big[\big(-(K-1)\overline{\bm{\eta}}^\top+\bm{1}^\top\big)\bm{\ell}\big(\bmg(X)\big)\big]\\
&=\mathbb{E}_{X\sim M}\big[-(K-1)\overline{\bm{\eta}}^\top \bm{\ell}\big(\bmg(X)\big)+\bm{1}^\top\bm{\ell}\big(\bmg(X)\big)\big]\\
&=\mathbb{E}_{(X,\overline{Y})\sim\overline{\mathcal{D}}}\big[-(K-1)\cdot\ell\big(\overline{Y},\bmg(X)\big)\big]\\
&\qquad\qquad\qquad+\bm{1}^\top\mathbb{E}_{X\sim M}\big[\bm{\ell}\big(\bmg(X)\big)\big]\\
&=\sum^K_{k=1}\overline{\pi}_k \cdot\mathbb{E}_{X\sim \overline{P}_k}\Big[ -(K-1)\cdot\ell\big(k,\bmg(X)\big)\\
&\qquad\qquad\qquad\quad\quad\quad+\bm{1}^\top\bm{\ell}\big(\bmg(X)\big) \Big]\\
&=\overline{R}(g;\overline{\ell})
\end{align*}
for the complementary loss,
$\overline{\ell}(k,\bmg):= -(K-1)\ell(k,\bmg) + \bm{1}^\top \bm{\ell}(\bmg)$,
which concludes the proof.
\end{proof}

\section{Proof of Corollary \ref{corollary}}\label{sec:corollary2proof}
\begin{proof}
\begin{align*}
\overline{R}(\bmg;\overline{\ell})
&=\mathbb{E}_{\overline{D}}[\overline{\ell}(\overline{Y},\bmg(X))]\\
&=\mathbb{E}_{\overline{D}}[-(K-1)\ell(\overline{Y},\bmg(X))+\sum^K_{j=1}\ell(j,\bmg(X))]\\
&=\mathbb{E}_{\overline{D}}\big[-(K-1)[M_2-\overline{\ell}(\overline{Y},\bmg(X))]+M_1\big]\\
&=(K-1)\mathbb{E}_{\overline{D}}[\overline{\ell}(\overline{Y},\bmg(X))]+M_1-(K-1)M_2\\
&=(K-1)\mathbb{E}_{\overline{D}}[\overline{\ell}(\overline{Y},\bmg(X))]-M_1+M_2
\end{align*}
The second equality holds because we use \eqref{scalar_complementary_loss}.  The third equality holds because we are using losses that satisfy $\sum_j\ell(j,\bmg(x))=M_1$ for all $x$ and $\ell(\overline{y},\bmg(x))+\overline{\ell}(\overline{y},\bmg(x))=M_2$ for all $x$ and $\overline{y}$.  The 4th equality rearranges terms.  The 5th equality holds because $M_1-(K-1)M_2=-M_1+M_2$ for $\overline{\ell}_\text{OVA}$ and $\overline{\ell}_\text{PC}$.  This can be easily shown by using $M_1=K$ and $M_2=2$ for $\overline{\ell}_{\text{OVA}}$, and $M_1=K(K-1)/2$ and $M_2=K-1$ for $\overline{\ell}_{\text{PC}}$.
\end{proof}

\begin{table}[]
  \center
  \small
  \caption{Summary statistics of benchmark datasets.  In the experiments with validation dataset in Section \ref{sec:validation_experiments}, train data is further splitted into train/validation with a ratio of 9:1. Fashion is Fashion-MNIST and Kuzushiji is Kuzushiji-MNIST.}
  \tabcolsep=0.07cm
    \begin{tabular}{l|cc|cccc|c}\toprule
    Name & \# Train & \# Test & \# Dim & \# Classes & Model\\\midrule
    MNIST &60k&10k&784& 10 & Linear, MLP\\\midrule
    Fashion &60k&10k&784& 10 & Linear, MLP\\\midrule
    Kuzushiji &60k&10k&784& 10 & Linear, MLP\\\midrule
    CIFAR-10 &50k&10k&2,048& 10 & DenseNet, Resnet\\\bottomrule
  \end{tabular}
  \label{tb:summary_datasets}
\end{table}

\section{Datasets}
\label{sec:datasets}
In the experiments in Section \ref{sc:experiments}, we use 4 benchmark datasets explained below.
The summary statistics of the four datasets are given in Table \ref{tb:summary_datasets}.
\begin{itemize}
\item MNIST\footnote{\url{http://yann.lecun.com/exdb/mnist/}} \citep{Lecun98gradient-basedlearning} is a 10 class dataset of handwritten digits: $1,2\ldots, 9$ and $0$.  Each sample is a $28\times28$ grayscale image.
\item Fashion-MNIST\footnote{\url{https://github.com/zalandoresearch/fashion-mnist}} \citep{fashion} is a 10 class dataset of fashion items: T-shirt/top, Trouser, Pullover, Dress, Coat, Sandal, Shirt, Sneaker, Bag, and Ankle boot.  Each sample is a $28\times28$ grayscale image.
\item Kuzushiji-MNIST\footnote{\url{https://github.com/rois-codh/kmnist}} \citep{clanuwat18neurips} is a 10 class dataset of cursive Japanese (``Kuzushiji'') characters.  Each sample is a $28\times28$ grayscale image.
\item CIFAR-10\footnote{\url{https://www.cs.toronto.edu/~kriz/cifar.html}} is a 10 class dataset of various objects: airplane, automobile, bird, cat, deer, dog, frog, horse, ship, and truck.  Each sample is a colored image in $32\times32\times3$ RGB format.  It is a subset of the 80 million tiny images dataset \citep{Torralba08pami}.
\end{itemize}

\newpage
\onecolumn
\section*{Errata (Nov. 19, 2019)}

We have found errors in our previous implementation for the forward method \citep{yu17eccv}, and would like to report the updated results based on the fixed implementation.

\begin{figure*}[h]
\vspace{-5mm}
\centering
\subfigure[MNIST, linear]{\includegraphics[bb = 0 0 351 348, scale = 0.3339]{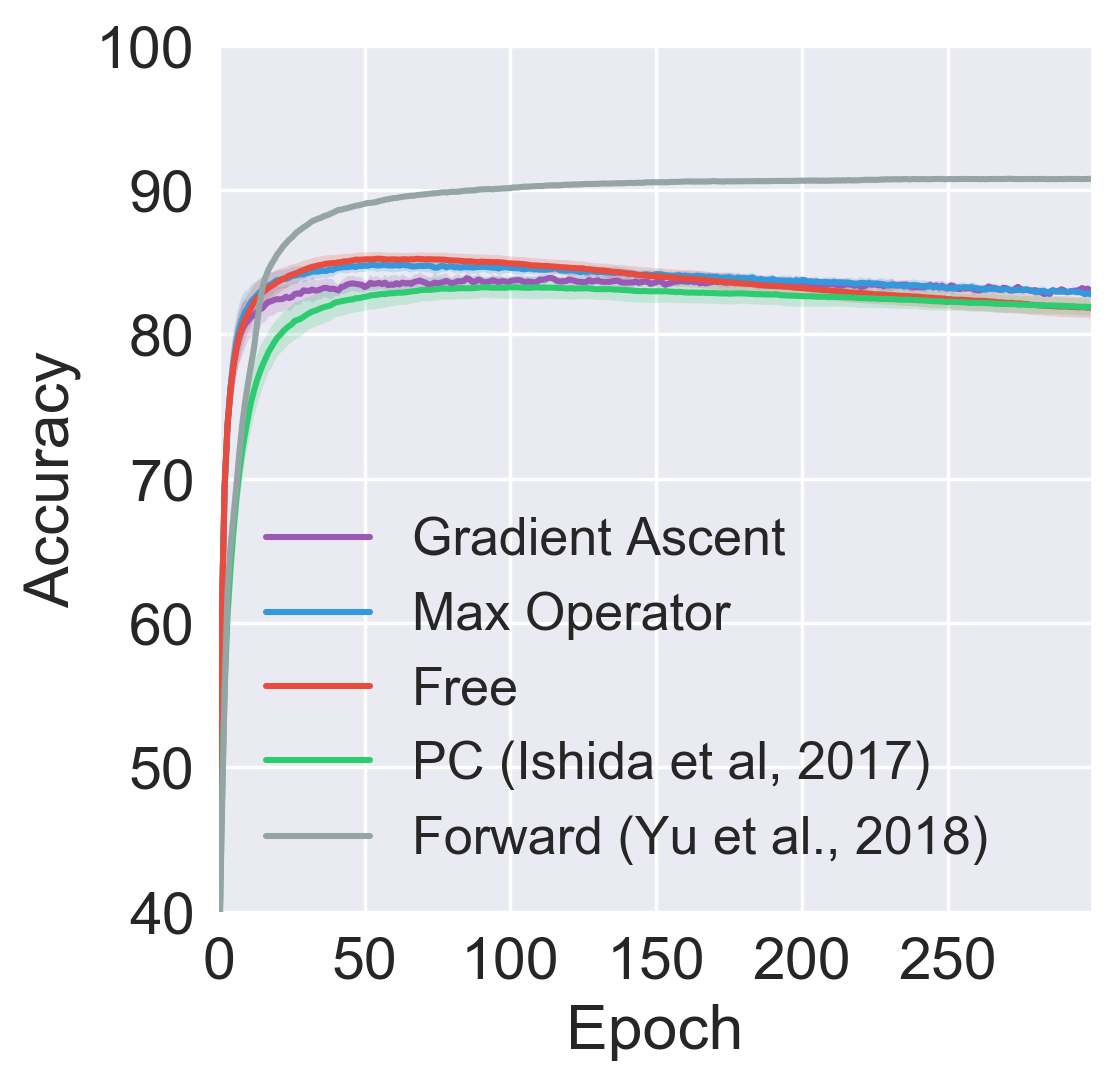}}
\hspace{0.001\textwidth}
\subfigure[MNIST, MLP]{\includegraphics[bb = 0 0 351 348, scale = 0.3339]{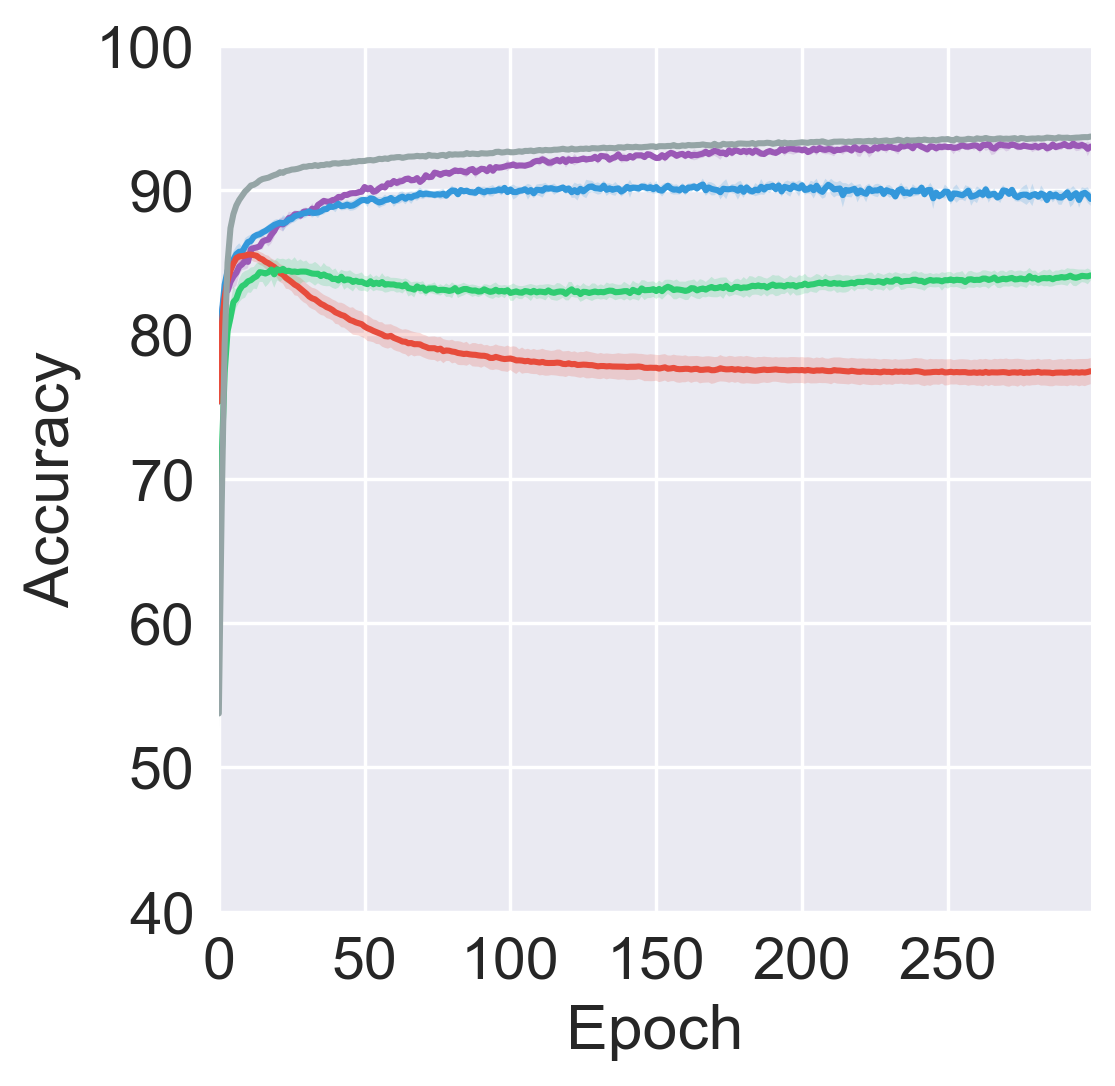}}
\hspace{0.001\textwidth}
\subfigure[Fashion-MNIST, linear]{\includegraphics[bb = 0 0 351 348, scale = 0.3339]{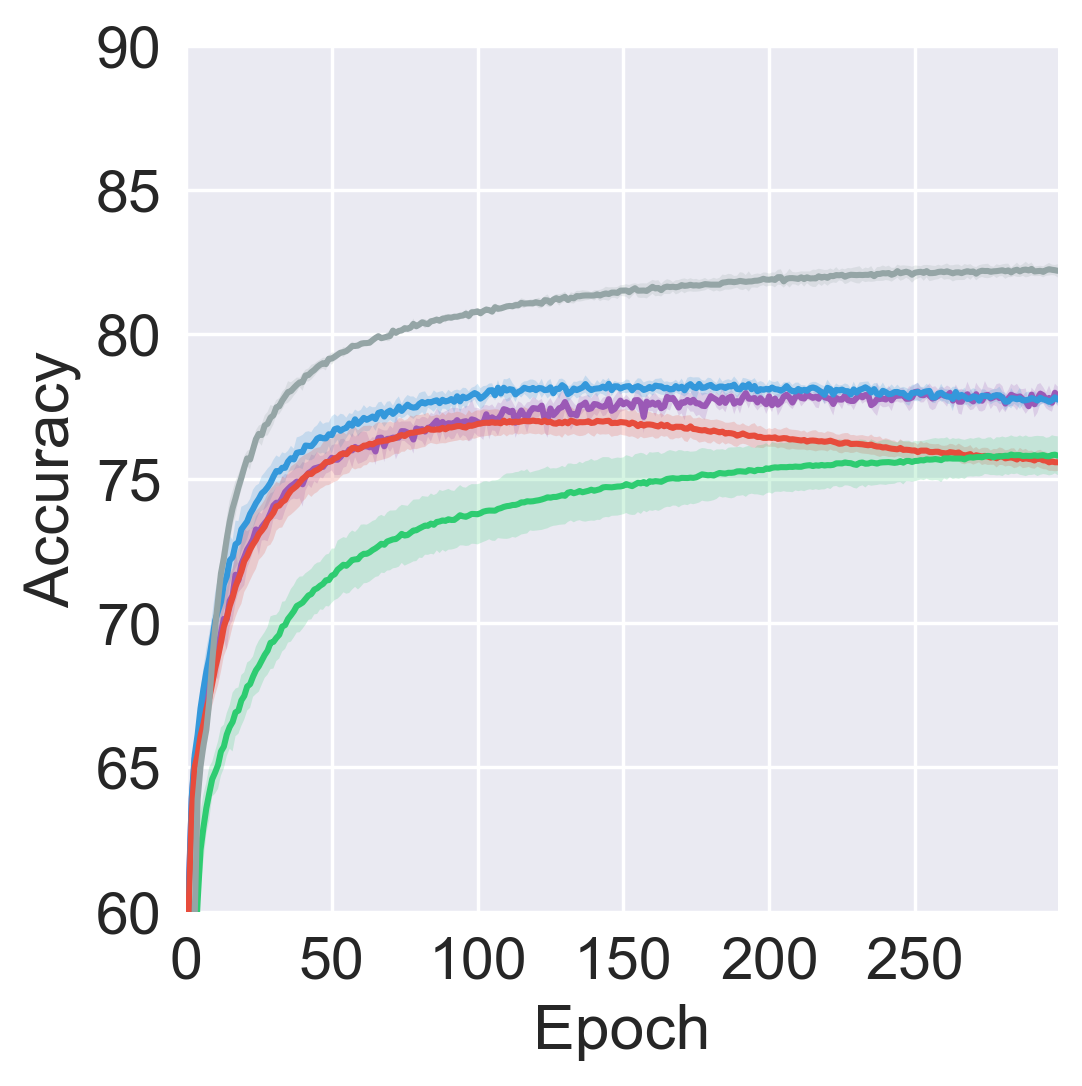}}
\hspace{0.001\textwidth}
\subfigure[Fashion-MNIST, MLP]{\includegraphics[bb = 0 0 351 348, scale = 0.3339]{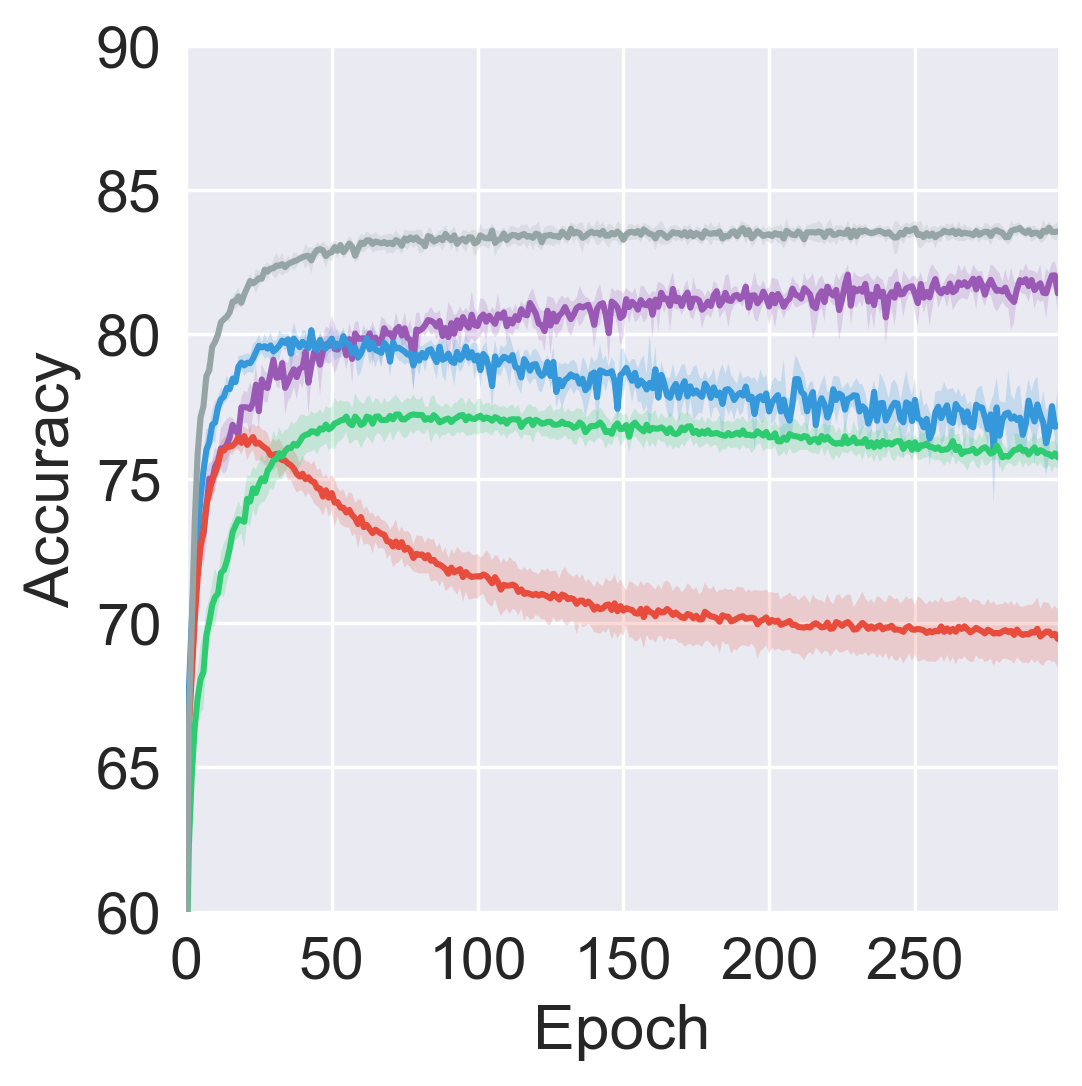}}\\
\vspace{-3mm}
\subfigure[Kuzushiji-MNIST, linear]{\includegraphics[bb = 0 0 351 348, scale = 0.3339]{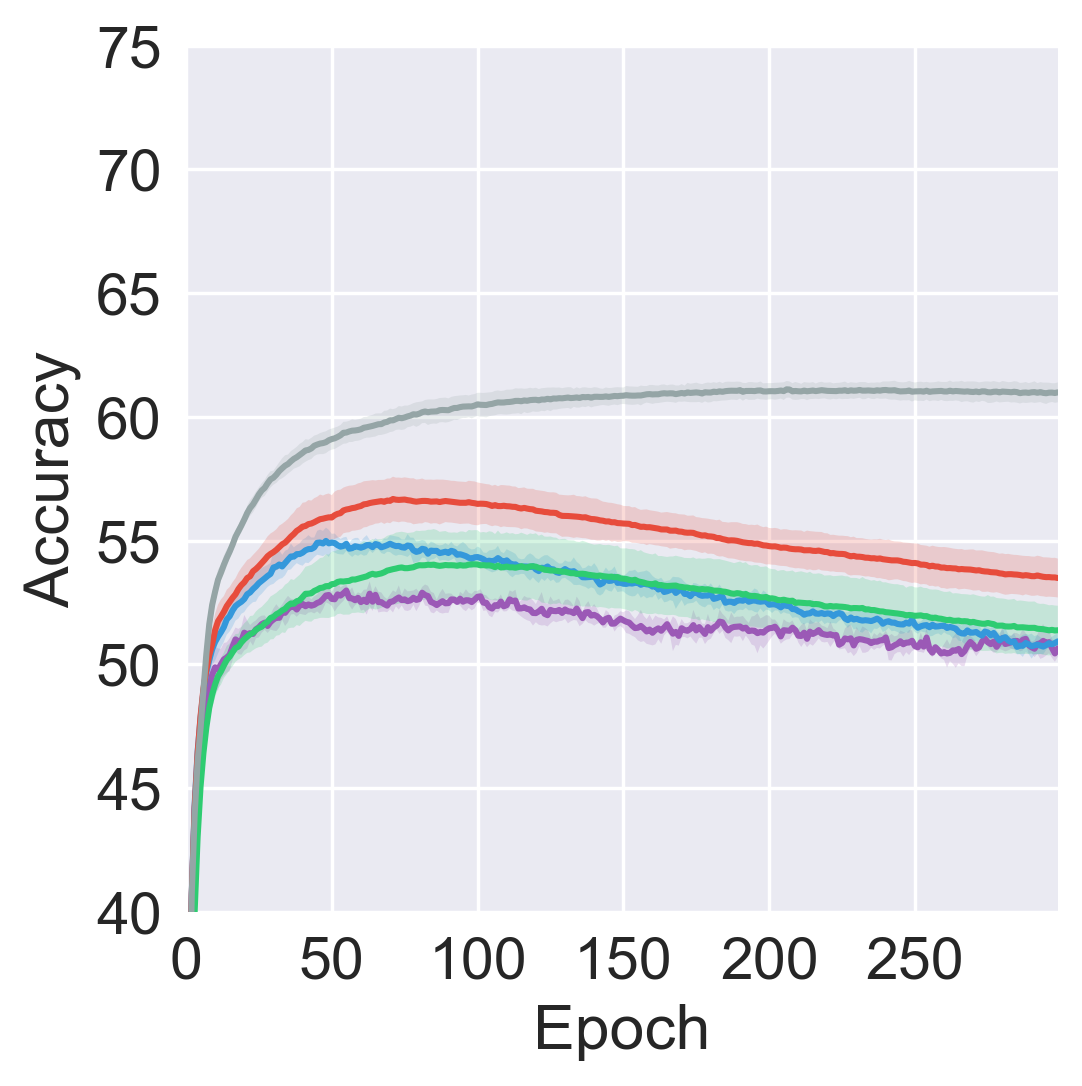}}
\hspace{0.001\textwidth}
\subfigure[Kuzushiji-MNIST, MLP]{\includegraphics[bb = 0 0 351 348, scale = 0.3339]{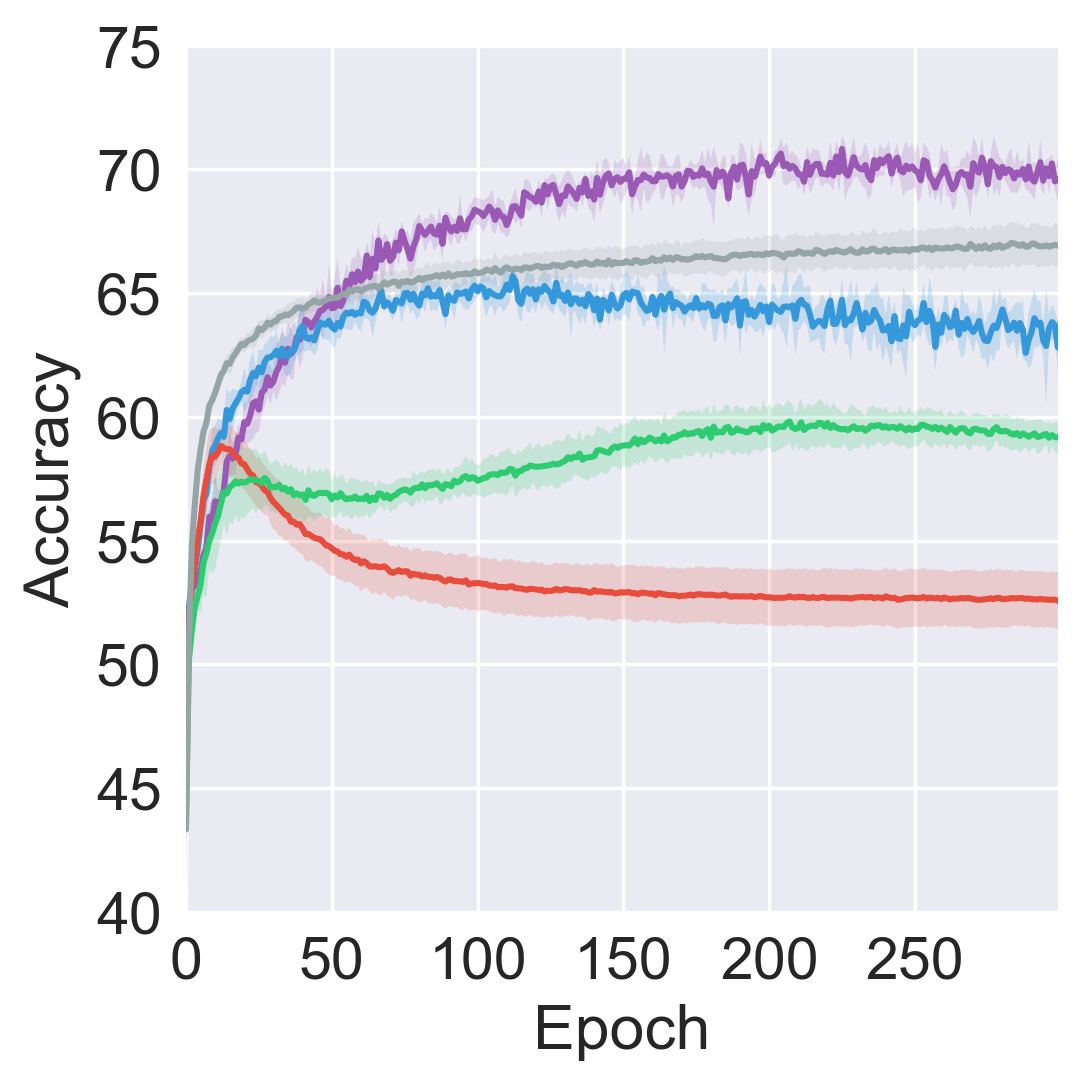}}
\hspace{0.001\textwidth}
\subfigure[CIFAR-10, DenseNet]{\includegraphics[bb = 0 0 351 348, scale = 0.3339]{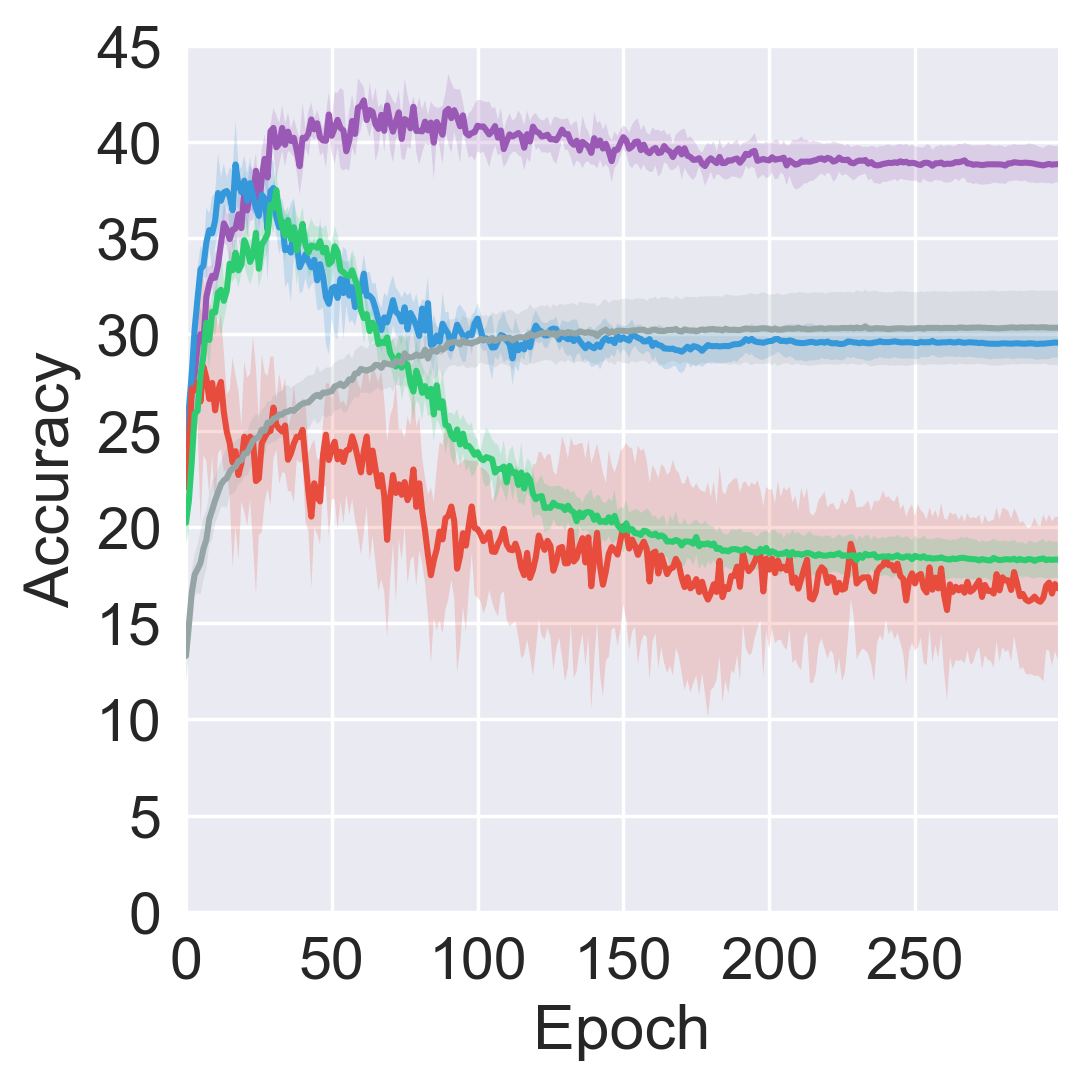}}
\hspace{0.001\textwidth}
\subfigure[CIFAR-10, ResNet]{\includegraphics[bb = 0 0 351 348, scale = 0.3339]{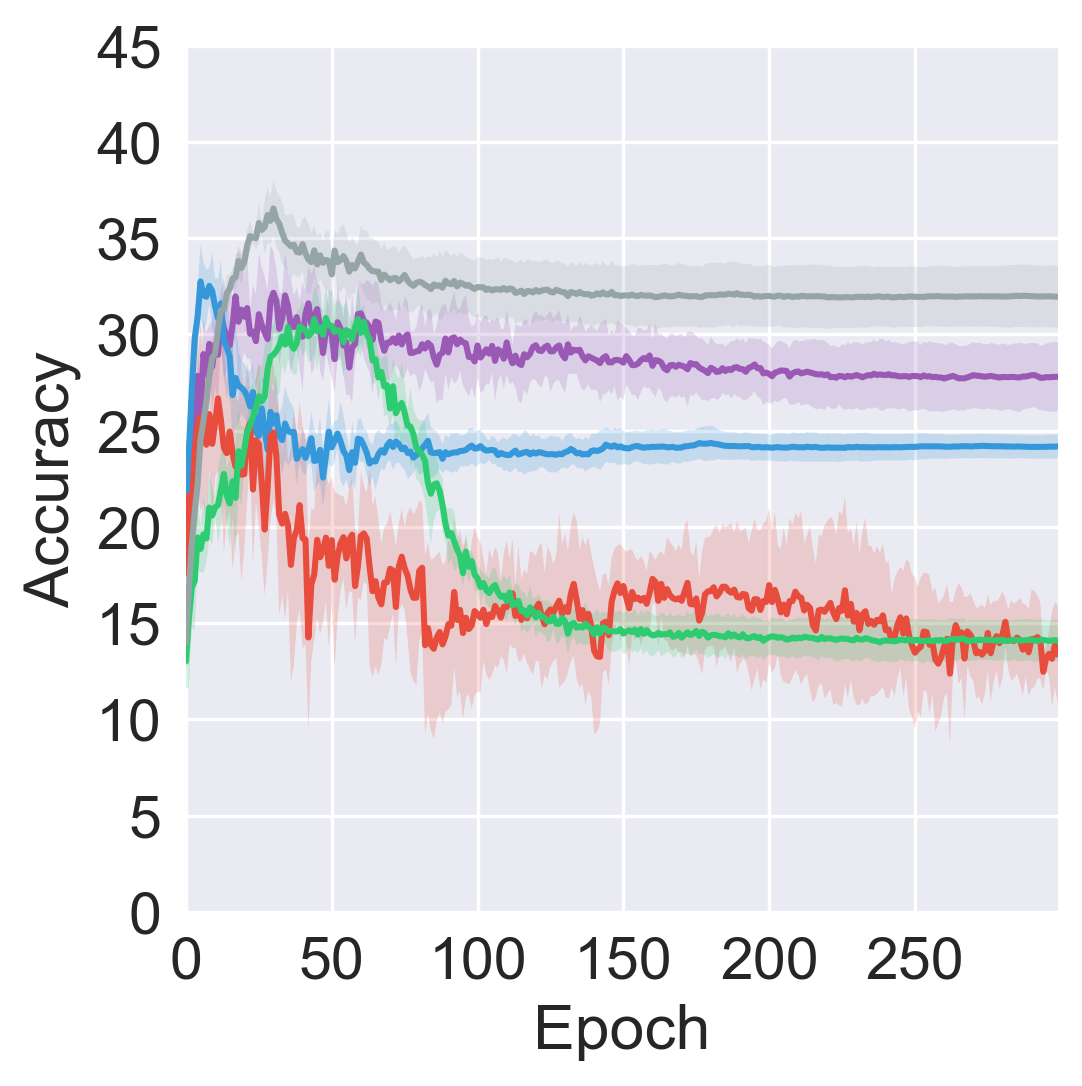}}
\caption{Updated results of Figure \ref{fig:results}.  The forward method has been swapped with the fixed implementation.}
\label{fig:updatedresults}
\end{figure*}

In Figure \ref{fig:updatedresults}, the forward method performs better than previously reported.  This is especially true for linear models.  For neural network models,  the results seem to be dataset-dependent:  For MNIST and Fashion-MNIST, the proposed gradient ascent method is similar to the forward method.  For Kuzushiji-MNIST, the proposed gradient ascent method is still better than the forward method.  For CIFAR-10, the proposed gradient ascent method with DenseNet still performs the best with around 40\% accuracy.  In Table \ref{tb:validation2}, the proposed gradient ascent method is better for CIFAR-10, but the forward method is better for MNIST and Fashion-MNIST.  The two methods perform similarly for Kuzushiji-MNIST.

\begin{figure*}[h]
\centering
  \centering
  \includegraphics[bb = 0 0 1287 802, scale = 0.52]{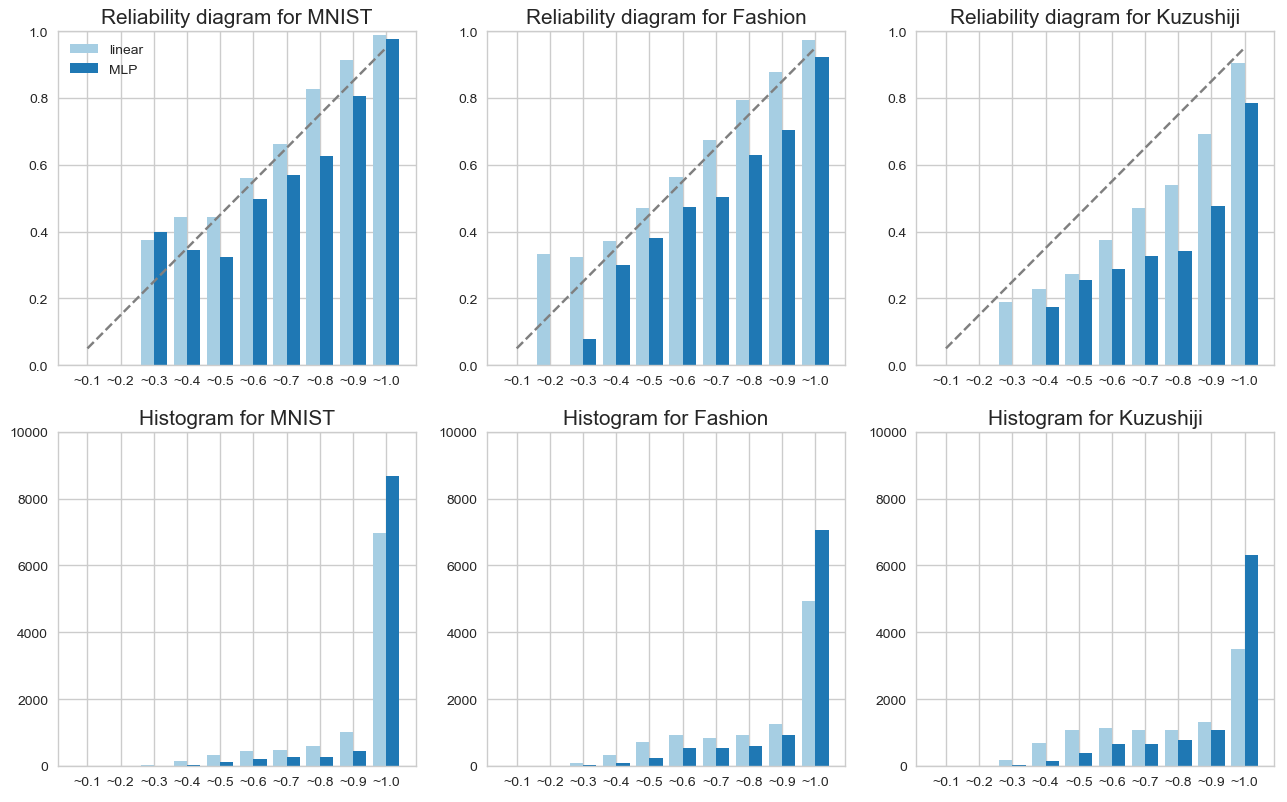}
  \caption{The bottom figures show the histogram of the output of the softmax layer in the forward method, with 10 bins in the horizontal axis, for MNIST, Fashion-MNIST, and Kuzushiji-MNIST.  The light blue color shows the linear model and the dark blue color shows the MLP model.  The top figures show the reliability diagrams for the same datasets.  The vertical axis shows the proportion of correct predictions in each bins.  The gray dotted line shows the identity function as an ideal case.}
  \label{fig:conf}
\end{figure*}

Addtionally, we investigate the reason behind the good performance of forward methods with a linear model.  In Figure \ref{fig:conf}, we visualize the reliability diagrams  \cite{confcalibration} and histograms of the softmax output of the forward method, for MNIST, Fashion-MNIST, and Kuzushiji-MNIST.  We can see that the linear model is much more confidence-calibrated compared to MLP models.  The forward method requires the model to be flexible in order to guarantee that the solution gives the true class posterior under the clean joint distribution, given uncountably infinite training data.  In the figures, however, we can see that with finite training data, a flexible model can be over-confident (further away from the gray dotted line), while a linear model is more confidence-calibrated (more closer to the gray dotted line).

\begin{table*}[h] 
  \center
  \small
  \caption{Updated results of Table \ref{tb:validation}.  The results in the 4th and 6th columns have been swapped with the fixed implementation.}
  \label{tb:validation2}
  \tabcolsep=0.2cm
    \begin{tabular}{l|ccc|cc}\toprule
    Dataset & \emph{GA (Free)} & \emph{PC (PC)} & \emph{Fwd (Fwd)} & \emph{PC (Free)} & \emph{Fwd (Free)} \\\midrule
    MNIST&$88.1\pm2.5\%$&$79.3\pm3.3\%$&$\bm{\underline{93.3\pm0.2\%}}$&$80.2\pm2.9\%$&$92.2\pm0.6\%$\\\midrule
    Fashion&$78.7\pm1.4\%$&$74.7\pm1.6\%$&$\bm{\underline{82.7\pm0.4\%}}$&$75.7\pm1.2\%$&$82.5\pm0.6\%$ \\\midrule
    Kuzushiji&$63.8\pm1.1\%$&$56.7\pm4.9\%$&$\bm{\underline{64.1\pm0.4\%}}$&$56.1\pm4.2\%$&$63.9\pm1.9\%$ \\\midrule
    CIFAR-10&$\bm{\underline{36.8\pm0.6\%}}$&$33.4\pm2.0\%$&$33.2\pm0.9\%$&$25.9\pm7.6\%$&$33.7\pm1.1\%$ \\\bottomrule
  \end{tabular}
\end{table*}

\end{document}